\documentclass[11pt]{article}

\usepackage[final]{acl}

\usepackage{times}
\usepackage{latexsym}
\usepackage{amssymb} 

\usepackage[T1]{fontenc}

\usepackage[utf8]{inputenc}

\usepackage{microtype}

\usepackage{inconsolata}

\usepackage{graphicx}
\usepackage{booktabs}
\usepackage{multirow}
\usepackage[table]{xcolor} 
\usepackage{amsfonts}
\usepackage{algpseudocode}
\usepackage{algorithm}
\usepackage{amsmath} 
\usepackage{pgfplots}
\usepackage{bbm}
\usepackage[most]{tcolorbox}

\usepackage{pifont}
\newcommand{\cmark}{{\color{green!60!black}\ding{51}}} 
\newcommand{\xmark}{{\color{red}\ding{55}}}            

\title{Beyond Outcome Verification: Verifiable Process Reward Models for Structured Reasoning 
}

\author{
 \textbf{Massimiliano Pronesti\textsuperscript{1,2}},
 \textbf{Anya Belz\textsuperscript{2}},
 \textbf{Yufang Hou\textsuperscript{1,3}}
\\
 \textsuperscript{1}IBM Research Europe - Ireland,
 \textsuperscript{2}Dublin City University,\\
 \textsuperscript{3}IT:U Interdisciplinary Transformation University Austria
\\
 \small{
   \textbf{Correspondence:} \href{mailto:massimiliano.pronesti@ibm.com}{massimiliano.pronesti@ibm.com}, \href{mailto:yufang.hou@it-u.at}{yufang.hou@it-u.at}
 }
}

\begin{document}
\maketitle
\begin{abstract}
Recent work on reinforcement learning with verifiable rewards (RLVR) has shown that large language models (LLMs) can be substantially improved using outcome-level verification signals, such as unit tests for code or exact-match checks for mathematics. In parallel, process supervision has long been explored as a way to shape the intermediate reasoning behaviour of LLMs, but existing approaches rely on neural judges to score chain-of-thought steps, leaving them vulnerable to opacity, bias, and reward hacking.
To address this gap, we introduce \emph{Verifiable Process Reward Models} (VPRMs), a reinforcement-learning framework in which intermediate reasoning steps are checked by deterministic, rule-based verifiers. We apply VPRMs to risk-of-bias assessment for medical evidence synthesis, a domain where guideline-defined criteria and rule-based decision paths enable programmatic verification of reasoning traces.
Across multiple datasets, we find that VPRMs generate reasoning that adheres closely to domain rules and achieve substantially higher coherence between step-level decisions and final labels. 
Results show that VPRMs achieve up to 20\% higher F1 than state-of-the-art models and 6.5\% higher than verifiable outcome rewards, with substantial gains in evidence grounding and logical coherence.
\end{abstract}

\begin{figure*}
    \centering
    \includegraphics[width=.75\linewidth, height=0.3\textheight]{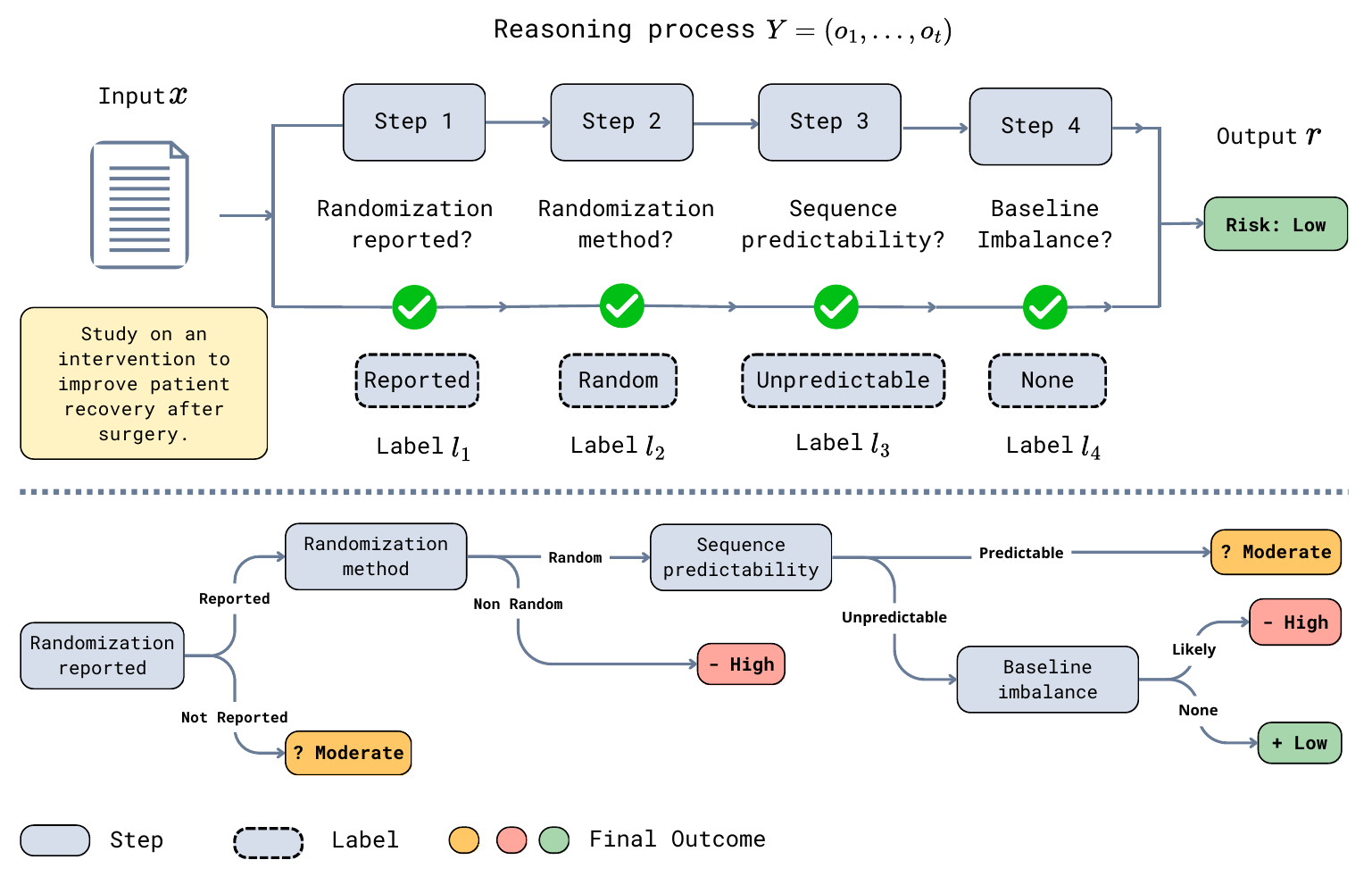}
    \caption{Illustration of the verifiable reasoning setup for risk-of-bias assessment (type A: bias arising from the randomisation process). 
Top: given an input study $x$, the model produces a structured reasoning trace 
$Y=(o_1,\ldots,o_T)$ with step-level labels $(\ell_1,\ell_2,\ell_3,\ell_4)$, each corresponding 
to a guideline-defined assessment question.
Bottom: the corresponding rule-based decision tree, which deterministically maps 
each combination of step-level labels to low (\textbf{+}), high (\textbf{-}), or moderate (\textbf{?}) risk.}

    \label{fig:task}
\end{figure*}

\section{Introduction}  

Large language models (LLMs) have made remarkable progress in complex natural language processing tasks, including reasoning, planning, and structured decision making~\cite{brown2020language, openai_o1_2024, openai_gpt5_2025}.
Reinforcement learning with verifiable rewards (RLVR) has recently emerged as a robust alternative to preference-based reinforcement learning, enabling LLMs to improve using reward signals derived from deterministic checks such as program test suites or exact-match mathematical evaluation~\cite{deepseekr1, lambert2025tulu}. By grounding supervision in objective verifiers rather than learned reward models, RLVR avoids many of the alignment failures associated with neural reward hacking and has produced state-of-the-art performance in code generation and mathematical reasoning~\cite{wang2025reinforcement,zhang-zuo-2025-grpo, deepseekr1, yang2025qwen3technicalreport}.  

However, outcome-only RLVR provides rewards solely at the terminal step of reasoning, offering no guarantees about whether the model followed a valid intermediate process. To address this limitation, several extensions augment RLVR with structural or auxiliary signals, such as masked-and-reordered self-supervision~\cite{wang2025masked} or self-verification mechanisms~\cite{zeng2025simplerlzoo}. These works strengthen RLVR but still operate fundamentally at the level of \textit{outcome verification}.
Most importantly, none of the above methods provide a fully verifiable form of process supervision, and existing approaches that score intermediate Chain-of-Thought (CoT) steps rely on neural judges~\cite{lightman2024lets,zhang-etal-2025-lessons, zou2025reasonfluxprm} which reintroduce opacity, bias, and opportunities for reward hacking~\cite{amodei2016concrete, skalse2022defining}. 

To date, no work has demonstrated that process rewards themselves can be made verifiable on tasks whose structure admits deterministic symbolic checking. Yet such a capability would be highly desirable: if intermediate reasoning steps can be validated explicitly, then reinforcement learning could optimise not only for correct outcomes, but also for transparent, logically sound reasoning. Crucially, verifiable step-level rewards would eliminate the aforementioned problems associated with neural process rewards.


This leaves a key open problem: can reinforcement learning be used to train models whose entire reasoning trajectory is rewarded only when each intermediate step satisfies domain-defined, rule-based criteria?
To address this gap, we introduce \textbf{Verifiable Process Reward Models} (VPRMs), a reinforcement-learning framework in which each reasoning step is assessed by a deterministic verifier grounded in explicit task guidelines. VPRMs provide fine-grained, step-level reward signals that complement outcome-level verification, guiding optimisation toward reasoning traces that are both correct and aligned with domain logic. Crucially, we prove that under mild assumptions, VPRMs offer theoretical guarantees that gradient-based updates assign positive expected weight to correct reasoning trajectories and negative weight to inconsistent ones, thereby encouraging sound reasoning.

To evaluate this framework, we consider a challenging, real-world structured-reasoning task: risk-of-bias (RoB) assessment in clinical systematic reviews. 
In this setting, studies must be evaluated for susceptibility to systematic error~\cite{cochranehandbook}, and domain guidelines prescribe a rigid sequence of reasoning steps and decision rules that make the task uniquely amenable to verifiable process supervision. Figure~\ref{fig:task} illustrates the verifiable reasoning process for assessing randomisation bias, 
one of the RoB domains defined in the Cochrane RoB tool for randomised trials \cite{sterne2019rob2}. 

Across multiple models and RoB domains, we compare VPRMs against outcome-only RLVR, neural process-reward baselines, and pretrained LLMs prompted for Chain-of-Thought (CoT). Our results show that VPRM-trained models achieve substantially higher accuracy and 
more coherent reasoning traces, showing that verifiable process supervision offers a more reliable optimization signal for both result correctness and process soundness. 

In summary, our contributions are as follows: (i) we propose a verifiable process reward framework that integrates deterministic step-level verification with reinforcement learning for dense, interpretable supervision over reasoning trajectories (Section~\ref{sec:vprm}); (ii) we show that, under mild reward-separation assumptions, verifiable process rewards encourage correct reasoning (Section~\ref{subsec:reward-separation}, Appendix~\ref{app:theo}); and (iii) we validate the approach on risk-of-bias assessment in medical systematic reviews, demonstrating significant improvements over outcome-only and neural process-reward baselines (Section~\ref{sec:results}).


\section{Preliminaries}

\subsection{Policy Optimisation Algorithms}
\paragraph{Group Relative Policy Optimization (GRPO)} GRPO~\cite{deepseek-math} is a widely-adopted group-based reinforcement learning method that optimises a policy by comparing multiple candidate completions for the same input.

For each passage $\mathbf{x}$, $G$ candidate completions $\{y_i\}_{i=1}^G \sim \pi_\text{old}( \cdot \mid \mathbf{x})$ are sampled from the reference policy $\pi_\text{old}$ to encourage robustness and diversity. These completions are scored using a reward model. The raw rewards $R_i$ are then normalised across the group:
{
	\begin{align*}
A_i = \frac{R_i - \mathbb{E}[R_j]}{\sqrt{\mathbb{V}[R_j]}}, \quad j \in \{i,...,G\}
	\end{align*}
}
where $\mathbb{E}[R_j]$ and $\mathbb{V}[R_j]$ are respectively the mean and variance of the rewards for the group of responses. The policy is optimised using a clipped, KL-regularised objective that encourages agreement with high-reward behaviours while maintaining proximity to a reference model $\pi_{\mathrm{ref}}$:

{\footnotesize
	\begin{align*}
	\mathcal{L}_{\text{GRPO}}(\theta) =\;&
		\mathbb{E} \bigg[ \frac{1}{G} \sum_{i=1}^G \frac{1}{|y_i|} \sum_{t=1}^{|y_i|} 
		\min
        \Big( p_{i,t}(\theta) A_i,\,\\
		&\quad \hspace{-1cm}
		\text{clip}(p_{i,t}(\theta), 1-\varepsilon,1+\varepsilon) A_i \Big) \notag  - \beta\, \text{KL}[\pi_\theta \,\|\, \pi_{\text{ref}}] \bigg]
	\end{align*}
}


\noindent where $\beta$ governs the regularisation strength and  $p_{i,t}(\theta)$ is the token-level probability ratio defined as follows: 
\[
p_{i,t}(\theta) = \frac{\pi_\theta (y_{i,t} \mid x, y_{i,<t})}{\pi_{\theta_\text{old}} (y_{i,t} \mid x, y_{i,<t})}
\]

\paragraph{Dynamic sAmpling Policy Optimization (DAPO)} Building on GRPO, DAPO~\cite{yu2025dapo} removes the KL penalty, introduces a clip-higher mechanism, incorporates dynamic sampling, applies a token-level policy gradient loss, and adopts overlong reward shaping.  The key improvement is dynamic sampling, by over-sampling and filtering out prompts with the accuracy equal to 1 and 0, leaving all prompts in the batch with effective gradients, avoiding dampening the gradient signals for
model training with a larger variance in the gradient. This leads to the following maximisation objective:

{\footnotesize
	\begin{align*}
	\mathcal{L}_{\text{DAPO}}(\theta) =\;&
		\mathbb{E} \bigg[ \frac{1}{\sum^G_{i=1} |y_i|} \sum_{i=1}^G \sum_{t=1}^{|y_i|} 
		\min
        \Big( p_{i,t}(\theta) A_i,\,\\
		&\quad \hspace{-1cm}
		\text{clip}(p_{i,t}(\theta), 1-\varepsilon_H,1+\varepsilon_L) A_i \Big) \notag \bigg],
        \\ 
        &\text{s.t.} \ 0 < |\{ y_i | \ \text{is\_equivalent}(y_i, a) \}| < G
	\end{align*}
}

\noindent This ensures that for the same input, the sampled set contains both correct and incorrect answers.

\subsection{Rule-based Reward Modeling}  In rule-based reward modeling, the reward signal is defined by explicit, hand-crafted rules that verify whether a model output satisfies task-specific constraints. This approach is particularly effective for verifiable tasks with clear notions of correctness, such as mathematical problem solving, program synthesis, or logical reasoning. The reward is computed deterministically by a verifier and does not rely on learned preference models. In its simplest form, the reward is binary:
\[
R(y) =
\begin{cases}
1 & \text{if the output is verified as correct}, \\
0 & \text{otherwise.}
\end{cases}
\]
This setting provides scalable, reliable supervision with minimal ambiguity.

\subsection{Systematic Reviews}
Systematic reviews provide a principled framework for aggregating empirical evidence through predefined search strategies, explicit inclusion criteria, and reproducible synthesis pipelines~\cite{cochranehandbook}. Their value lies in reducing subjective judgment in evidence collection and enabling structured comparison across heterogeneous studies. 
One of the gold-standard repositories of systematic reviews is the Cochrane library \cite{}, 
which contains curated reviews adhering to strict evidence-synthesis and bias-assessment protocols.


\subsection{Risk of Bias Assessment}
Risk of bias assessment is a core component of systematic reviews, providing structured evaluations of methodological flaws in primary studies that directly inform evidence synthesis and the credibility of review conclusions.
Assessments are structured into a fixed set of domains of bias, focusing on different aspects of trial design, conduct and reporting \cite{cochranehandbook, sterne2019rob2}. Within each domain, information about features of the trial that are relevant to risk of bias is collected and mapped to judgments of low, medium or high risk. 
This domain-based assessment identifies issues such as selection bias, measurement bias, and selective reporting, enabling subsequent evidence synthesis to appropriately weight studies not only in terms of their results, but also in terms of their risk of bias. Additional details on the different bias domains and the corresponding assessment reasoning process are provided in Appendix~\ref{app:rob}. 


\section{Verifiable Process Reward Models}
\label{sec:vprm}

\begin{figure*}
    \centering
    \includegraphics[height=0.26
    \textheight]{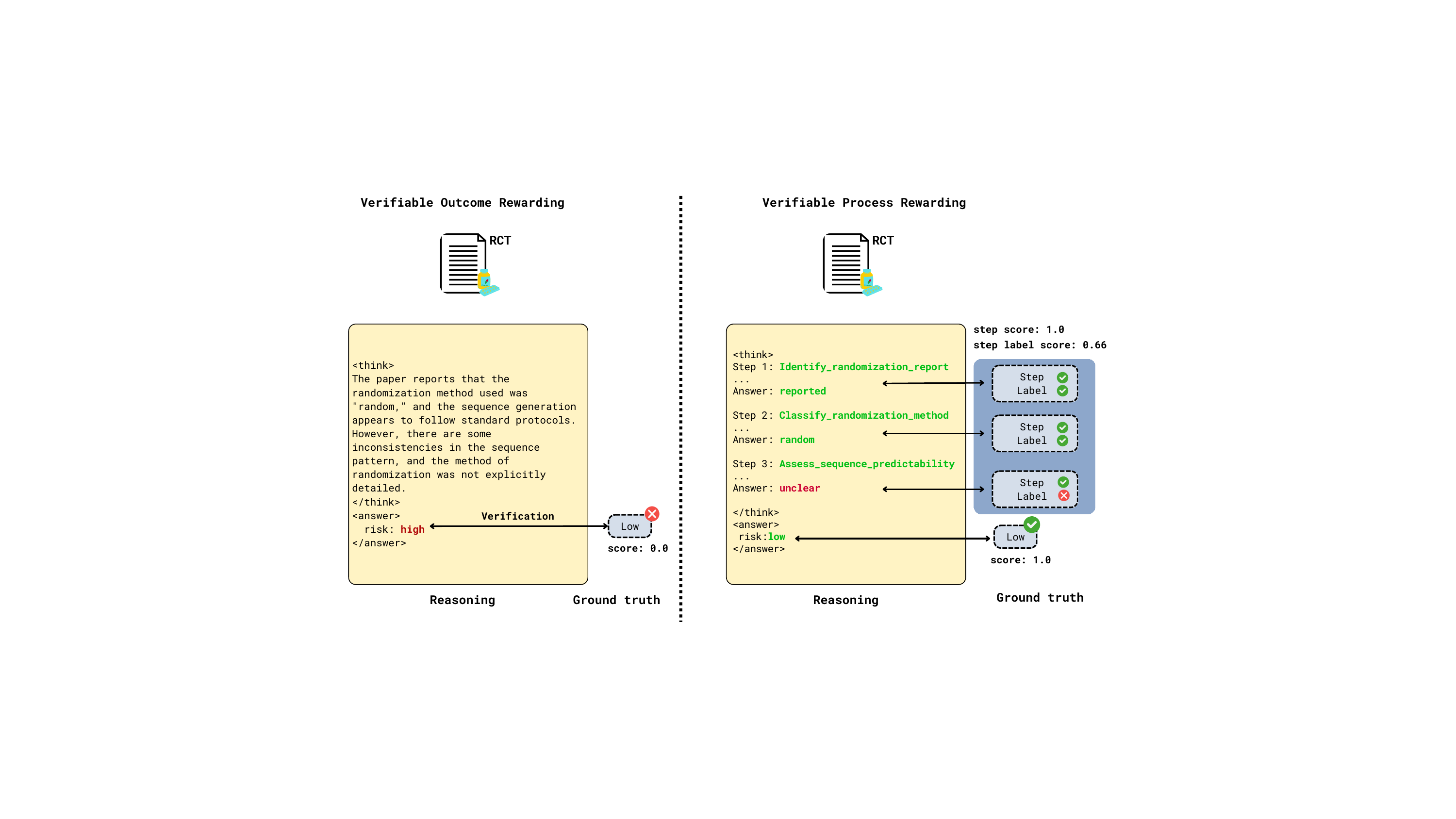}
\caption{Comparison between verifiable outcome rewards (left), which evaluates only the final risk label, and verifiable process rewards (right), which additionally verifies 
each reasoning step and its associated label.
}
    \label{fig:method}
\end{figure*}

We introduce \emph{Verifiable Process Reward Models} (VPRMs), a framework for process supervision in which intermediate reasoning steps are evaluated by deterministic, externally checkable verifiers rather than learned neural judges. The framework is presented in the context of risk assessment tasks, which naturally admit structured reasoning steps, discrete labels, and rule-based transitions that can be validated against domain guidelines. 

\subsection{Reasoning Trajectories and Steps}



For an input $x$, the model produces a reasoning trajectory
$Y = (o_1,\dots,o_T)$ with a stochastic policy
\[
\pi_\theta(Y \mid x) = \prod_{t=1}^T \pi_\theta(o_t \mid o_{<t}, x).
\]

Each step $t$ contains two discrete outputs:
(i) a step identifier $s_t \in \mathcal{S}$
and (ii) a step label $\hat\ell_t \in \mathcal{L}_t$, which represents the model’s answer for that step.
Domain guidelines specify, for each prefix $Y_{\le t}$, the \emph{gold} step
identifier $s_t^\star$ and gold step label $\ell_t^\star$ obtained by applying
the rule-based logic of the task.

\subsection{Verifiers and Process Rewards}



%

Correctness is evaluated by two bounded scoring functions:
\[
s_t^{\text{n}}(s_t, s_t^\star), \qquad
s_t^{\text{l}}(\hat\ell_t, \ell_t^\star),
\]
each mapping a model output and its corresponding gold value into $[0,1]$.
These provide a positive reward signal when the model selects the correct step
identifier and the correct label according to the task rules.
The instanteneous step-level verifiable reward is then defined as:
\[
r_t(Y;x)
= w_t^{\text{n}}\, s_t^{\text{n}}(s_t, s_t^\star)
+ w_t^{\text{l}}\, s_t^{\text{l}}(\hat\ell_t, \ell_t^\star),
\]
where $w_t^{\text{n}}, w_t^{\text{l}} \ge 0$ are preset weights. 

A terminal outcome reward $r_{\mathrm{label}}$ evaluates whether the final risk value predicted from the full reasoning trace matches the gold risk value. The full verifiable process reward is then
\[
R(Y;x) = \sum_{t=1}^T r_t (Y; x) + r_{\mathrm{label}}
\]

As illustrated in Figure~\ref{fig:method},  this reward is fully computable using deterministic, rule-based checks,
making all components of the reasoning trajectory verifiable.  

\subsection{Reward Separation and Optimisation Guarantee}
\label{subsec:reward-separation}



An interesting consequence of combining VPRMs with GRPO or DAPO is that the
resulting optimisation dynamics exhibit a clear structure: rule-consistent
trajectories are, in expectation, pushed in a beneficial direction.

Let $R(Y)$ denote the verifiable process reward assigned to a response $y$, and let $G$ be the number of responses sampled for the input $x$.  Let $\mathcal C$ be the event that $Y$ is \emph{correct} according to the
rule-based task semantics. Define the conditional expectations
\[
\mu_c := \mathbb E[R(Y)\mid \mathcal C], \qquad
\mu_i := \mathbb E[R(Y)\mid \mathcal C^c].
\]

We assume the following mild conditions (see
Appendix~\ref{app:theo} for the full formal statement and discussion):
(i) $R(Y)$ has finite variance,  
(ii) correct reasoning chains receive strictly larger expected reward than incorrect ones $(\mu_c > \mu_i)$, and  
(iii) a sufficiently large $G$ to ensure stable gradient updates. 

\begin{tcolorbox}[
	colback=gray!3,
	colframe=black!55,
	boxrule=0.35pt,
	arc=1.5pt,
	left=3pt,
	right=3pt,
	top=3pt,
	bottom=3pt
	]
	\textbf{Theorem 1.}
	Under the above hypotheses, the expected GRPO and DAPO advantage $\mathbb{E}[\hat A(Y)]
	$ satisfies
	\[
	\mathbb{E}[\hat A(Y)\mid \mathcal C] > 0,
	\qquad
	\mathbb{E}[\hat A(Y)\mid \mathcal C^c] < 0.
	\]
	Thus, both GRPO and DAPO assign positive expected weight to correct reasoning
	trajectories and negative weight to incorrect ones, raising the likelihood of
	correct reasoning in expectation.
\end{tcolorbox}

This follows from the theoretical results presented by \citet{rlvr-incentivizes}. A proof for both GRPO and DAPO objectives is provided in Appendix~\ref{app:theo}.

\section{Experiments}\label{sec:results}
\subsection{Training Dataset Creation}\label{sec:dataset-creation}
The first stage of our methodology involved the acquisition of a high-quality, human-aligned corpus suitable for training reward models. 

To this end, we build on the \textsc{CochraneForest}~\cite{pronesti-etal-2025-query} and \textsc{CochraneForestExt}~\cite{pronesti-etal-2025-enhancing} datasets, which provide two essential components for our task: (i) the forest plots extracted from Cochrane systematic reviews, and (ii) the \emph{full-text} primary studies corresponding to every trial included in those plots.  
The inclusion of full papers is critical, as risk-related signals often depend on methodological details available only in the complete manuscripts.

From these corpora, we retain exclusively the forest plots that contain an associated risk-of-bias map.  
Each such plot establishes an explicit correspondence between its set of included studies and their study-level bias assessments.  
We therefore define a single instance as a \emph{paper-risk pair} consisting of a full-text study and its aligned risk-of-bias definition and label extracted from the map.
The resulting dataset comprises 2,946 instances drawn from 104 systematic reviews, totalling 4M tokens.

\subsection{Synthetic Data Annotation for Structural Reasoning Processes}
\label{sec:syntheticdata}
Following the methodology described by \citet{pronesti-etal-2025-enhancing}, we enrich our dataset with step-level labels for RL using LlaMa 3.1 405B~\cite{grattafiori2024llama} with the
system prompt shown in Figure~\ref{fig:synth_prompt} (Appendix), temperature of 0.7 and 2,048 tokens generation limit.
An example data instance is provided in Table~\ref{tab:data_example} (Appendix); a human verification of the generated annotations in Appendix~\ref{app:silver_labels}. 

\subsection{Experimental Setup}
\paragraph{Training and Evaluation Datasets.}
For 
training, we use the corpus constructed with the methods from Section~\ref{sec:dataset-creation}, allocating 2{,}651 instances for training and 295 for validation. We also include the 774 training instances from the RoBBR Cochrane split~\cite{wang-etal-2025-measuring}. All training data are augmented with step-level labels (Section~\ref{sec:syntheticdata}). 

For evaluation, we consider three datasets.  
The first is \textsc{CochraneForest}~\cite{pronesti-etal-2025-enhancing}, which contains 1{,}846 instances drawn from 48 Cochrane Systematic Reviews and 202 forest plots.  
The second consists of the two test sets from the RoBBR benchmark:  
RoBBR Cochrane, which contains 906 datapoints originating from 204 papers included in 58 Cochrane reviews; and RoBBR–Non-Cochrane, which contains 2{,}489 datapoints drawn from 496 non-Cochrane reviews that collectively assess 496 papers. Dataset statistics are summarised in Table~\ref{tab:dataset_stats}, while per-risk-type statistics are reported in Table~\ref{tab:dataset_stats_per_risk} (Appendix).

\begin{table}[t]
\scalebox{0.73}{
\centering
\begin{tabular}{lcccccc}
\toprule
\textbf{Dataset} & \textbf{Train} & \textbf{Test} & \textbf{Total}  & \textbf{Avg tokens}\\
\midrule
\textsc{CochraneForestExt} & 2651 & 295 & 2946  & 13,596.9 \\
\textsc{CochraneForest} & – &  1846 & 1846 &  12,722.8 \\
RoBBR Cochrane & 774 & 906  & 1680  &  9,084.6 \\
RoBBR Non-Cochrane & – & 2489 &  2489 &  7,940.7\\
\bottomrule
\end{tabular}
}
\caption{Datasets statistics. Train/test split only applies to \textsc{CochraneForestExt} and RoBBR Cochrane. \textsc{CochraneForest} and RoBBR Non-Cochrane are used for testing.}
\label{tab:dataset_stats}
\end{table}

\paragraph{Evaluation Metrics.} We evaluate all models on the main prediction task using Accuracy and macro–F1, computed over the discrete risk labels.

In addition, for analyses (Section~\ref{sec:analyses}), we report \emph{Coherence} for the VPRM-trained models, defined as the proportion of datapoints for which the model’s predicted risk is consistent with the conclusion implied by its own intermediate reasoning. Formally, let $\hat r_i\in\mathcal R$ denote the final risk value predicted by the model for datapoint $i$, and let $\hat\ell_{i,1},\dots,\hat\ell_{i,T}$ denote the sequence of step-level labels produced along the corresponding reasoning trace. Let $D:\mathcal L_1\times\cdots\times\mathcal L_T\to\mathcal R$ be a fixed, externally specified decision function mapping step-level labels to a risk value. In our setting, this is the set of macros used in the RoB2 tool~\cite{sterne2019rob2} (See Figure~\ref{fig:task}). The coherence indicator for datapoint $i$ is then
\[
C_i := \mathbbm{1}\!\left\{\,\hat r_i = D(\hat\ell_{i,1},\dots,\hat\ell_{i,T})\,\right\}
\]
and the dataset-level Coherence is given by
\[
\mathrm{Coherence} := \frac{1}{N}\sum_{i=1}^N C_i
\]
By construction, Coherence measures the degree to which the model’s final conclusions are internally consistent with the reasoning signals expressed in its own intermediate steps.

\paragraph{Training Setup.} We conduct our training using a compact instruct models of recent release: \texttt{Qwen2.5-7B}~\cite{qwen2.5}.  We study two methodological regimes: supervised fine-tuning (SFT) with reasoning traces augmentation and reinforcement learning (RL) with verifiable rewards.  
SFT is conducted for $5$ epochs with a per-device batch size of $1$, a learning rate of $5\times 10^{-5}$, and the AdamW optimiser~\cite{adamw}.  
For RL, we investigate two policy-optimisation algorithms, \textsc{GRPO}~\cite{deepseek-math} and \textsc{DAPO}~\cite{yu2025dapo}, combined with two reward types: verifiable outcome reward and our verifiable process reward approach.  
All RL configurations are trained for $3$ epochs with a learning rate of $1\times 10^{-6}$, per-device batch size $1$, and $16$ sampled generations per batch.  
Further implementation details are provided in Appendix~\ref{app:hyp_and_apis}.

\begin{table*}[t]
	\small
	\centering
	\renewcommand{\arraystretch}{1.24}
	\setlength{\tabcolsep}{4pt}
	\begin{tabular}{lc|cc|cc|cc}
		\toprule
		\multirow{2}{*}{\textbf{Model}} & \multirow{2}{*}{\textbf{Think}} 
		& \multicolumn{2}{c|}{\cellcolor{gray!20}\textsc{\textbf{CochraneForest}}} 
		& \multicolumn{2}{c|}{\cellcolor{gray!20}\textbf{RoBBR Cochrane}}
		& \multicolumn{2}{c}{\cellcolor{gray!20}\textbf{RoBBR Non-Cochrane}} \\
		& & Acc & F1 & Acc & F1 & Acc & F1 \\
		\toprule

		\multicolumn{2}{c}{\cellcolor{gray!20}\textbf{Pretrained LLMs}} 
		& \multicolumn{2}{c|}{\cellcolor{gray!20}} 
		& \multicolumn{2}{c|}{\cellcolor{gray!20}} 
		& \multicolumn{2}{c}{\cellcolor{gray!20}} \\

		\texttt{GPT-4-0125} & \xmark & 52.4 & 41.6 & 56.0 & 47.9 & 47.8 & 42.3\\
        \texttt{GPT-OSS-20B} & \cmark & 61.4 & 43.9& 56.4 & 50.3 & 46.3 & 42.8 \\
        \texttt{GPT-OSS-120B} & \cmark & 67.1 & 49.8 & 59.5 & 51.0 & 48.8 & 44.2\\
		\texttt{Qwen2.5-7B} & \xmark & 32.9 & 31.6 & 35.8 & 34.1 & 36.4 & 34.5 \\
		\texttt{Qwen2.5-14B} & \xmark & 39.0 & 35.1 & 37.0 & 35.5 & 35.4 & 32.5  \\
		\texttt{Qwen2.5-72B} & \xmark & 51.3 & 42.1 & 56.1 & 51.0 & 47.5 & 43.6 \\
		\texttt{Llama-3.1-8B} & \xmark & 36.4 & 30.6 & 34.5 & 32.1 & 36.4 & 32.5 \\
		\texttt{Llama-3.1-70B} & \xmark & 38.8 & 30.2 & 49.5 & 40.0 & 42.5 & 38.9 \\
		\texttt{Llama-3.1-405B} & \xmark & 68.4 & 45.5 & 59.4 & 44.0 & 52.5 & 39.8 \\
		\texttt{DeepSeek-Qwen-7B} & \cmark & -- & -- & -- & -- & -- & -- \\
		\texttt{DeepSeek-Qwen-14B} & \cmark & 33.3 & 19.2 & 35.8 & 23.5 & 35.4 & 23.5 \\
		\texttt{DeepSeek-Qwen-32B} & \cmark & 40.8 & 35.9 & 44.9 & 40.4 & 46.4 & 41.3 \\
        \texttt{DeepSeek-Llama-8B} & \cmark  & -- & -- & -- & -- & -- & -- \\
		\texttt{DeepSeek-Llama-70B} & \cmark & 44.2 & 33.5 & 57.3 & 41.2 & 48.3 & 42.7 \\
       \texttt{Granite-3.1-3B} & \xmark & 24.4 & 23.6 & 22.2 & 21.8 & 13.7 & 14.9 \\
       \texttt{Granite-3.1-8B} & \xmark & 24.7 & 22.0 & 35.8 & 31.6 & 33.2 & 28.2 \\ 
       \texttt{Granite-4.0-h-small (32B)} & \xmark & 48.2 & 33.5 & 45.4 & 41.2 & 40.9 & 33.1\\  
		\toprule
		\multicolumn{2}{c}{\cellcolor{gray!20}\textbf{Our Models}} 
		& \multicolumn{2}{c|}{\cellcolor{gray!20}} 
		& \multicolumn{2}{c|}{\cellcolor{gray!20}} 
		& \multicolumn{2}{c}{\cellcolor{gray!20}} \\

		\texttt{Qwen2.5-7B-SFT} & \cmark & 45.1 & 36.9 & 38.6 &   32.4  &  38.3 &  31.9 \\
		\texttt{Qwen2.5-7B-GRPO}  & \cmark & \underline{81.5} & \underline{70.2} & \underline{63.1} & \underline{58.0} & \underline{56.8} & \underline{45.1}\\
        \texttt{Qwen2.5-7B-DAPO}  & \cmark & 76.8 & 57.3 & 60.2 & 45.4 & 55.8 & 43.6 \\
        \texttt{Qwen2.5-7B-GRPO-VPRM}  & \cmark & \textbf{87.9} & \textbf{76.7} & \textbf{65.2} & \textbf{58.5} & \textbf{60.7} & \textbf{47.2} \\
        \texttt{Qwen2.5-7B-DAPO-VPRM}  & \cmark & 79.2 & 60.6 & 60.7 & 48.9 & 57.1 & 45.3 \\  
        
		\bottomrule
	\end{tabular}
	\caption{
    Evaluation results across models on three datasets, reporting Accuracy and macro-F1. ``--'' denotes unparsable or inconclusive outputs. Best results are bolded; second-best are underlined.
    }
	\label{tab:results}
\end{table*}

\paragraph{Model Baselines.} To validate our results, we
compare a range of open- and closed-source models, with and without reasoning capabilities. models are evaluated in zero-shot settings with
prompt and hyperparameters shown in Appendix~\ref{app:prompts}.
We include three main model families: Qwen
2.5~\cite{qwen2.5}, Llama 3.1~\cite{grattafiori2024llama}, and Granite 3.1~\cite{granite3}. In addition, we benchmark the distilled Qwen and Llama models derived from DeepSeek-R1~\cite{deepseekr1}. Lastly, we include one closed-source and two open-source models from OpenAI~\cite{openai2025gptoss120bgptoss20bmodel}.

To contextualise the effectiveness of our verifiable reward formulation, we also evaluate neural process-reward baselines. At present, no pretrained PRM exists for risk-of-bias assessment or, more broadly, for non-mathematical scientific reasoning.  Therefore, to approximate a general-purpose PRM, we follow prior work on using LLMs as step-level judges, prompting a model to assign correctness scores to each reasoning step~\cite{song-etal-2025-prmbench}. This setup has been shown to deliver competitive process-level feedback in domains where explicit PRM training data is unavailable, and therefore serves as a reasonable baseline for comparison. In addition, we train a policy using MedPRM~\cite{medprm} as a reward model, one of the first open-source PRMs for general medical reasoning.

\subsection{Main Results}
\paragraph{Comparison with Pretrained Baselines.} Table~\ref{tab:results} presents a performance comparison of pretrained and fine-tuned language models on the \textsc{CochraneForest} and RoBBR benchmarks. Across all datasets, models trained with verifiable rewards substantially outperform pretrained models, including large reasoning-enabled systems. Reinforcement learning with verifiable outcome rewards already yields strong gains over supervised fine-tuning, while incorporating verifiable process rewards consistently leads to further improvements in both accuracy and macro-F1. On \textsc{CochraneForest}, \texttt{Qwen2.5-7B} trained with VPRM achieves the best overall performance, and similar improvements are observed on both RoBBR Cochrane and Non-Cochrane. The latter result indicates that the benefits of verifiable process supervision generalise beyond the training distribution, rather than exploiting dataset-specific regularities.

\paragraph{Comparison with Neural PRMs.} Table~\ref{tab:prm_vs_vprm} compares verifiable process rewards against neural process-reward baselines. In all settings, models also receive the same 
verifiable outcome reward; the comparison isolates only the effect of the 
process-level supervision. While neural PRMs substantially improve over outcome-only training, they are consistently outperformed by VPRM. This performance gap suggests that learned step-level judges introduce noise and misalignment that limit their effectiveness, whereas deterministic, guideline-based verification provides a cleaner and more reliable optimisation signal. These results support the central claim that verifiable process rewards offer a stronger and more robust alternative to neural process supervision for complex, structured reasoning tasks.

\begin{table}[ht]
\small
	\centering
    \renewcommand{\arraystretch}{1}
	\begin{tabular}{l|cc}
		\toprule
		\textbf{Method} & \textbf{Acc} & \textbf{F1} \\
		\midrule
		\rowcolor{gray!15}
		\multicolumn{3}{l}{\textbf{Neural PRMs}} \\
		\texttt{Qwen2.5-7B-GRPO-PRM-GPT-OSS}   & 78.2 & 56.1 \\
		\texttt{Qwen2.5-7B-GRPO-MedPRM}  &  76.8  &  53.4     \\
		\midrule
		\rowcolor{gray!15}
		\multicolumn{3}{l}{\textbf{Verifiable Rewards}} \\
		\texttt{Qwen2.5-7B}   &   32.9     & 31.6       \\
		\texttt{Qwen2.5-7B-GRPO}   & 81.5 & 70.2            \\
		\texttt{Qwen2.5-7B-GRPO-VPRM}  & \textbf{87.9} & \textbf{76.7}         \\
		\bottomrule
	\end{tabular}
	\caption{Performance comparison between neural judges, rule-based rewarding and verifiable process rewarding on \textsc{CochraneForest}.}
	\label{tab:prm_vs_vprm}
\end{table}

\subsection{Ablation Studies} To assess the contribution of different components of our verifiable reward formulation, 
we conduct two ablation studies: 
the structure of process supervision and the inclusion of an outcome-level reward. For process supervision, we compare a \emph{steps-only} reward that verifies whether the model follows the correct sequence of reasoning steps, irrespective of the correctness of their content, against the full VPRM, which additionally evaluates the correctness of each step and enforces consistency with the guideline-defined decision structure. For outcome supervision, we train each variant both with and without a verifiable outcome reward.

Table~\ref{tab:ablation} shows that removing the outcome reward leads to substantial performance degradation, indicating that step-structure verification alone is insufficient to reliably optimize the task. Nevertheless, even in this setting, the full VPRM outperforms the steps-only variant, demonstrating the importance of verifying not just the presence but also the correctness and logical composition of intermediate reasoning steps. When the outcome reward is included, performance improves markedly, and the full VPRM consistently achieves the best results, showing that combining verifiable outcome supervision with fine-grained, correctness-aware process rewards yields the strongest learning signal.

\begin{table}[t]
\small
\centering
\renewcommand{\arraystretch}{1}
\begin{tabular}{l|cc}
\toprule
\textbf{Setting} & \textbf{Acc} & \textbf{F1} \\
\midrule
\rowcolor{gray!15}
		\multicolumn{3}{l}{\textbf{w/o Outcome Reward}}\\
\midrule
\quad Steps-only process reward      & 34.4 & 32.3 \\
\quad Full VPRM                     & 40.2 & 35.3 \\
\midrule
\rowcolor{gray!15}
		\multicolumn{3}{l}{\textbf{w/\ Outcome Reward}}\\
\midrule
\quad Steps-only process reward      & 83.1 & 71.8 \\
\quad Full VPRM                     & 87.9 & 76.7 \\
\bottomrule
\end{tabular}
\caption{Ablation study on outcome and process reward components on \textsc{CochraneForest}. 
We report Accuracy (Acc) and F1; lower blocks compare models with and without outcome reward, 
and columns compare steps-only supervision to the full VPRM formulation.}
\label{tab:ablation}
\end{table}

\subsection{Analyses}\label{sec:analyses}
\paragraph{Impact of Thought Process.} Table~\ref{tab:coherence} reports Coherence on the \textsc{CochraneForest} testset for the models trained with VPRMs compared with pretrained baselines prompted to output process labels. In addition, we report \emph{Coherent Accuracy (CA)}, defined as the accuracy restricted to coherent instances. That is, among the datapoints for which the model’s final prediction is consistent with the decision implied by its own reasoning steps (i.e., $C_i = 1$), we measure the proportion whose final predicted label is also correct. Formally, letting $\hat{y}_i$ and $y_i$ denote the predicted and gold labels respectively, we define
\[
\mathrm{CA}
=
\frac{
\sum_{i=1}^{N} \mathbbm{1}\!\left[C_i = 1 \;\wedge\; \hat{y}_i = y_i\right]
}{
N
}
\]
CA therefore quantifies the reliability of the model’s predictions conditioned on coherence. 

Results show that pretrained models exhibit low coherence and very low CA, indicating that even when their step-level reasoning appears self-consistent, it rarely leads to correct final judgments. In contrast, VPRM-trained models achieve both substantially higher coherence and high CA, demonstrating that they not only follow the decision logic faithfully but also produce accurate conclusions when they do so, suggesting improved robustness and interpretability.

\begin{figure}
\centering
\resizebox{\columnwidth}{!}{\begin{tikzpicture}
  \begin{axis}[
    width=11cm,   
    height=6.6cm, 
    grid=both,
    xlabel={Step},
    ylabel={Training Reward},
    xmin=0, xmax=2000,
    ymin=0.1, ymax=1.02,
    legend style={font=\small, legend columns=3, at={(0.49,1.02)}, anchor=south}
  ]

  \addplot[green!80, mark=o, thick, smooth] coordinates {
    (1,0.1868489533662796)
    (20,0.3330729126930237)
    (40,0.4638020783662796)
    (60,0.5662760317325592)
    (80,0.63359375)
    (100,0.47200521528720857)
    (120,0.5489583253860474)
    (140,0.5643229097127914)
    (160,0.692578125)
    (180,0.7212239384651185)
    (200,0.6917968630790711)
    (220,0.5654947936534882)
    (240,0.6576822996139526)
    (260,0.7519531130790711)
    (280,0.5716145753860473)
    (300,0.664453125)
    (320,0.6950520753860474)
    (340,0.6765625)
    (360,0.5442708194255829)
    (380,0.6427083253860474)
    (400,0.7076822996139527)
    (420,0.8473958253860474)
    (440,0.6546875)
    (460,0.6332031309604644)
    (480,0.7868489503860474)
    (500,0.5783854335546493)
    (520,0.7427083253860474)
    (540,0.5317708432674408)
    (560,0.7739583373069763)
    (580,0.6850260317325592)
    (600,0.6571614444255829)
    (620,0.782421875)
    (640,0.817187488079071)
    (660,0.7942708432674408)
    (680,0.8548177003860473)
    (700,0.7248697757720948)
    (720,0.6324218869209289)
    (740,0.7102864503860473)
    (760,0.654296875)
    (780,0.7447916746139527)
    (800,0.8680989623069764)
    (820,0.7822916626930236)
    (840,0.6755208313465119)
    (860,0.7675781130790711)
    (880,0.8104166507720947)
    (900,0.8505208253860473)
    (920,0.646875011920929)
    (940,0.8145833373069763)
    (960,0.6976562380790711)
    (980,0.8359374880790711)
    (1000,0.786328113079071)
    (1020,0.7989583253860474)
    (1040,0.6213541746139526)
    (1060,0.7578125)
    (1080,0.794140613079071)
    (1100,0.7410156190395355)
    (1120,0.5985677182674408)
    (1140,0.732421875)
    (1160,0.6106770694255829)
    (1180,0.8710937261581421)
    (1200,0.7888020634651184)
    (1220,0.770703113079071)
    (1240,0.826171875)
    (1260,0.6908854007720947)
    (1280,0.8680989503860473)
    (1300,0.760937488079071)
    (1320,0.7343749880790711)
    (1340,0.7572916507720947)
    (1360,0.8440104007720948)
    (1380,0.8089843630790711)
    (1400,0.7626302003860473)
    (1420,0.7755208134651184)
    (1440,0.7123697698116302)
    (1460,0.8580729007720947)
    (1480,0.8348958134651184)
    (1500,0.6567708313465118)
    (1520,0.7562499880790711)
    (1540,0.8891927123069763)
    (1560,0.8979166507720947)
    (1580,0.8998697757720947)
    (1600,0.7363281130790711)
    (1620,0.9162760257720948)
    (1640,0.831249988079071)
    (1660,0.8010416507720948)
    (1680,0.8281249761581421)
    (1700,0.7109374761581421)
    (1720,0.7859374761581421)
    (1740,0.9026041507720948)
    (1760,0.7618489503860474)
    (1780,0.7778645753860474)
    (1800,0.7225260317325592)
    (1820,0.7579427003860474)
    (1840,0.8141927003860474)
    (1860,0.736328113079071)
    (1880,0.7794270575046539)
    (1900,0.8114583373069763)
    (1920,0.8411458134651184)
    (1940,0.8181770753860474)
    (1960,0.7115885257720947)
    (1980,0.8618489503860474)
    (2000,0.8131510138511657)
  };
  \addlegendentry{Process Reward}

  \addplot[cyan!70, mark=triangle*, thick, smooth] coordinates {
    (1,0.190476194024086)
    (20,0.3121434718370438)
    (40,0.37996031939983366)
    (60,0.46009221076965334)
    (80,0.6875)
    (100,0.7)
    (120,0.636800754070282)
    (140,0.6585317492485047)
    (160,0.6375)
    (180,0.634375)
    (200,0.7625)
    (220,0.7118951618671417)
    (240,0.625)
    (260,0.696875)
    (280,0.734375)
    (300,0.728125)
    (320,0.6375)
    (340,0.8492063522338867)
    (360,0.6523809552192688)
    (380,0.6595238097012043)
    (400,0.75)
    (420,0.85)
    (440,0.65)
    (460,0.75)
    (480,0.703125)
    (500,0.596875)
    (520,0.688125)
    (540,0.55)
    (560,0.69)
    (580,0.75)
    (600,0.648412698507309)
    (620,0.65)
    (640,0.75625)
    (660,0.8375)
    (680,0.7516129016876221)
    (700,0.75)
    (720,0.5524065554141998)
    (740,0.7742943525314331)
    (760,0.696875)
    (780,0.7438764035701751)
    (800,0.85)
    (820,0.759375)
    (840,0.653125)
    (860,0.65625)
    (880,0.640625)
    (900,0.8373015880584717)
    (920,0.6866567492485046)
    (940,0.85)
    (960,0.65)
    (980,0.74)
    (1000,0.8125)
    (1020,0.7)
    (1040,0.790625)
    (1060,0.7)
    (1080,0.640625)
    (1100,0.753125)
    (1120,0.6)
    (1140,0.646875)
    (1160,0.8)
    (1180,0.84375)
    (1200,0.7)
    (1220,0.9)
    (1240,0.76875)
    (1260,0.71)
    (1280,0.75)
    (1300,0.75)
    (1320,0.6833333373069763)
    (1340,0.875)
    (1360,0.784375)
    (1380,0.8625)
    (1400,0.95)
    (1420,0.95)
    (1440,0.85)
    (1460,0.85)
    (1480,0.65)
    (1500,0.809375)
    (1520,0.69375)
    (1540,0.85)
    (1560,0.903125)
    (1580,0.7)
    (1600,0.90625)
    (1620,0.85)
    (1640,0.8)
    (1660,0.846875)
    (1680,0.8)
    (1700,0.896875)
    (1720,0.7492063522338868)
    (1740,0.940625)
    (1760,0.8984127044677734)
    (1780,0.9079861164093017)
    (1800,0.75)
    (1820,0.8023809552192688)
    (1840,0.95)
    (1860,0.755606758594513)
    (1880,0.75)
    (1900,0.8015873074531555)
    (1920,0.74375)
    (1940,0.75)
    (1960,0.8032258033752442)
    (1980,0.85)
    (2000,0.79)
  };
  \addlegendentry{Accuracy Reward}

  \addplot[red!50, mark=x, thick, smooth] coordinates {
    (1,0.96875)
    (20,0.93125)
    (40,0.99375)
    (60,0.9625)
    (80,1.0)
    (100,0.984375)
    (120,0.990625)
    (140,0.990625)
    (160,1.0)
    (180,0.996875)
    (200,1.0)
    (220,0.99375)
    (240,0.996875)
    (260,1.0)
    (280,0.990625)
    (300,0.996875)
    (320,0.996875)
    (340,1.0)
    (360,1.0)
    (380,0.99375)
    (400,1.0)
    (420,1.0)
    (440,0.996875)
    (460,0.98125)
    (480,1.0)
    (500,0.990625)
    (520,1.0)
    (540,1.0)
    (560,1.0)
    (580,0.996875)
    (600,0.996875)
    (620,1.0)
    (640,1.0)
    (660,0.996875)
    (680,0.996875)
    (700,1.0)
    (720,0.990625)
    (740,0.990625)
    (760,1.0)
    (780,0.9875)
    (800,1.0)
    (820,0.996875)
    (840,0.996875)
    (860,1.0)
    (880,1.0)
    (900,0.99375)
    (920,0.996875)
    (940,1.0)
    (960,1.0)
    (980,1.0)
    (1000,1.0)
    (1020,1.0)
    (1040,0.9875)
    (1060,0.996875)
    (1080,1.0)
    (1100,1.0)
    (1120,1.0)
    (1140,1.0)
    (1160,1.0)
    (1180,1.0)
    (1200,1.0)
    (1220,0.996875)
    (1240,1.0)
    (1260,0.996875)
    (1280,1.0)
    (1300,1.0)
    (1320,1.0)
    (1340,1.0)
    (1360,0.996875)
    (1380,0.996875)
    (1400,1.0)
    (1420,1.0)
    (1440,1.0)
    (1460,1.0)
    (1480,1.0)
    (1500,1.0)
    (1520,1.0)
    (1540,0.996875)
    (1560,1.0)
    (1580,1.0)
    (1600,1.0)
    (1620,1.0)
    (1640,1.0)
    (1660,1.0)
    (1680,1.0)
    (1700,1.0)
    (1720,0.996875)
    (1740,1.0)
    (1760,0.99375)
    (1780,0.996875)
    (1800,1.0)
    (1820,0.996875)
    (1840,0.99375)
    (1860,0.990625)
    (1880,0.996875)
    (1900,1.0)
    (1920,1.0)
    (1940,1.0)
    (1960,0.99375)
    (1980,1.0)
    (2000,1.0)
  };
  \addlegendentry{Thought Format Reward}

  \end{axis}
\end{tikzpicture}}
\caption{
Reward dynamics. Format rewards plateau early, while accuracy and process rewards improve gradually, indicating that LLMs quickly learn structure but continue to refine quality.
}
\label{fig:reward_dynamics}
\end{figure}
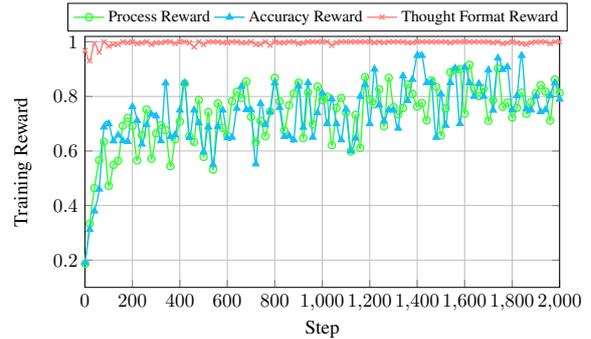

\paragraph{Reward Dynamics.} 
Figure~\ref{fig:reward_dynamics} shows that process and correctness rewards follow closely aligned trajectories: both rise sharply early, stabilize within the same oscillatory range, and peak at similar points. This alignment indicates that improved step-level reasoning directly improves final-label correctness. In contrast, the thought-format reward saturates quickly and remains flat, contributing little once formatting is learned. Overall, the strongly correlated shapes of the process and correctness curves highlight that VPRM training drives coherent, mutually reinforcing gains in intermediate and final reasoning behaviour.

\begin{table}[t]
\small
\centering
\renewcommand{\arraystretch}{0.94}
\begin{tabular}{l|c|c}
\toprule
\textbf{Model} & \textbf{Coherence} & \textbf{CA} \\
\midrule
\texttt{GPT-OSS-120B}& 36.2 & 28.5 \\
\texttt{Qwen2.5-72B} & 44.3 & 24.9 \\
\texttt{Llama-3.1-405B} & 50.7 & 27.1 \\
\midrule
\texttt{Qwen2.5-7B-GRPO-VPRM} & \textbf{89.5} & \textbf{75.0}   \\
\texttt{Qwen2.5-7B-DAPO-VPRM} & \underline{80.1} & \underline{69.4} \\
\bottomrule
\end{tabular}

\caption{Coherence scores for Qwen models trained with DAPO and GRPO using verifiable process rewards compared with pretrained LLMs on \textsc{CochraneForest}. Best results are bolded;  second-best underlined.}
\label{tab:coherence}
\end{table}

\section{Related Work}
For a more detailed overview of related work, refer
to Appendix~\ref{app:related_work}. Prior work extends reinforcement learning with verifiable rewards (RLVR) beyond outcome-only supervision by adding structural signals, such as masked-and-reordered self-supervision or self-verification modules that guide reasoning trajectories~\cite{wang2025masked, zeng2025simplerlzoo, rlvr-incentivizes}. While effective, these methods remain centered on terminal-outcome verification.
In parallel, process-supervision approaches score intermediate reasoning steps using neural judges~\cite{lightman2024lets,zhang-etal-2025-lessons, zou2025reasonfluxprm}, providing dense feedback but relying on non-verifiable model-based evaluations prone to bias and reward hacking~\cite{amodei2016concrete, skalse2022defining}. We bridge this gap by combining RLVR with step-wise rule-based verification, enabling transparent and verifiable process supervision.


\section{Conclusion}


In this paper, we introduce verifiable process rewards that integrate deterministic step-level verification with reinforcement learning, provide theoretical guarantees under mild assumptions, and demonstrate substantial empirical gains on risk-of-bias assessment in medical systematic reviews.

Our results indicate that verifiable process supervision is a practical and
robust approach to inducing reliable reasoning behaviour in large language
models, opening the door to broader applications in structured scientific
and decision-making tasks.
\section*{Limitations}
While VPRMs offer strong guarantees for structured reasoning tasks, several limitations remain. First, the approach relies on the existence of deterministic, domain-specific rules; tasks lacking well-defined
intermediate reasoning steps may not benefit directly. Second, our empirical
evaluation is currently focused on risk-of-bias assessment; generalisation
to other domains, particularly open-ended reasoning tasks, remains to be
established.  

Additionally, the approach assumes that the model can produce reasoning
traces in a format compatible with the verifiers; misalignment between model
output and verifier expectations could reduce reward effectiveness, especially in the context of smaller models. Finally,
while VPRMs reduce reliance on neural reward models, they do not fully
eliminate other sources of model bias or errors arising from incomplete
guidelines. Addressing these challenges will be critical for deploying
verifiable process supervision in broader, real-world applications.

\bibliography{custom}

\begin{thebibliography}{38}
\providecommand{\natexlab}[1]{#1}

\bibitem[{Amodei et~al.(2016)Amodei, Olah, Steinhardt, Christiano, Schulman, and Man{\'e}}]{amodei2016concrete}
Dario Amodei, Chris Olah, Jacob Steinhardt, Paul Christiano, John Schulman, and Dan Man{\'e}. 2016.
\newblock Concrete problems in ai safety.
\newblock \emph{arXiv preprint arXiv:1606.06565}.

\bibitem[{Brown et~al.(2020)Brown, Mann, Ryder, Subbiah, Kaplan, Dhariwal, Neelakantan, Shyam, Sastry, Askell et~al.}]{brown2020language}
Tom~B. Brown, Benjamin Mann, Nick Ryder, Melanie Subbiah, Jared Kaplan, Prafulla Dhariwal, Arvind Neelakantan, Pranav Shyam, Girish Sastry, Amanda Askell, and 1 others. 2020.
\newblock Language models are few‑shot learners.
\newblock \emph{Advances in Neural Information Processing Systems (NeurIPS)}, 33.

\bibitem[{Chandler et~al.(2019)Chandler, Cumpston, Li, Page, and Welch}]{cochranehandbook}
Jacqueline Chandler, Miranda Cumpston, Tianjing Li, Matthew~J. Page, and Vivian~A. Welch. 2019.
\newblock Cochrane handbook for systematic reviews of interventions.
\newblock \emph{Hoboken: Wiley}, 4.

\bibitem[{Dias et~al.(2025)Dias, Moreira, and Comba}]{dias2025robin}
Abel~Corr{\^e}a Dias, Viviane~Pereira Moreira, and Jo{\~a}o Luiz~Dihl Comba. 2025.
\newblock Robin: A transformer-based model for risk of bias inference with machine reading comprehension.
\newblock \emph{Journal of Biomedical Informatics}, 166:104819.

\bibitem[{Granite~Team(2024)}]{granite3}
IBM Granite~Team. 2024.
\newblock Granite 3.0 language models.
\newblock \emph{URL: https://github.com/ibm-granite/granite-3.0-language-models}.

\bibitem[{Grattafiori et~al.(2024)Grattafiori, Dubey, Jauhri, Pandey, Kadian, Al-Dahle, Letman, Mathur, Schelten, Vaughan et~al.}]{grattafiori2024llama}
Aaron Grattafiori, Abhimanyu Dubey, Abhinav Jauhri, Abhinav Pandey, Abhishek Kadian, Ahmad Al-Dahle, Aiesha Letman, Akhil Mathur, Alan Schelten, Alex Vaughan, and 1 others. 2024.
\newblock The {LL}a{M}a 3 herd of models.
\newblock \emph{arXiv preprint arXiv:2407.21783}.

\bibitem[{Guo et~al.(2025)Guo, Yang, Zhang, Song, Zhang, Xu, Zhu, Ma, Wang, Bi, Zhang, Yu, Wu, Wu, Gou, Shao, Li, Gao, Liu, Xue, Wang, Wu, Feng, Lu, Zhao, Deng, Zhang, Ruan, Dai, Chen, Ji, Li, Lin, Dai, Luo, Hao, Chen, Li, Zhang, Bao, Xu, Wang, Ding, Xin, Gao, Qu, Li, Guo, Li, Wang, Chen, Yuan, Qiu, Li, Cai, Ni, Liang, Chen, Dong, Hu, Gao, Guan, Huang, Yu, Wang, Zhang, Zhao, Wang, Zhang, Xu, Xia, Zhang, Zhang, Tang, Li, Wang, Li, Tian, Huang, Zhang, Wang, Chen, Du, Ge, Zhang, Pan, Wang, Chen, Jin, Chen, Lu, Zhou, Chen, Ye, Wang, Yu, Zhou, Pan, Li, Zhou, Wu, Ye, Yun, Pei, Sun, Wang, Zeng, Zhao, Liu, Liang, Gao, Yu, Zhang, Xiao, An, Liu, Wang, Chen, Nie, Cheng, Liu, Xie, Liu, Yang, Li, Su, Lin, Li, Jin, Shen, Chen, Sun, Wang, Song, Zhou, Wang, Shan, Li, Wang, Wei, Zhang, Xu, Li, Zhao, Sun, Wang, Yu, Zhang, Shi, Xiong, He, Piao, Wang, Tan, Ma, Liu, Guo, Ou, Wang, Gong, Zou, He, Xiong, Luo, You, Liu, Zhou, Zhu, Xu, Huang, Li, Zheng, Zhu, Ma, Tang, Zha, Yan, Ren, Ren, Sha, Fu, Xu, Xie, Zhang, Hao, Ma, Yan, Wu, Gu,
  Zhu, Liu, Li, Xie, Song, Pan, Huang, Xu, Zhang, and Zhang}]{deepseekr1}
Daya Guo, Dejian Yang, Haowei Zhang, Junxiao Song, Ruoyu Zhang, Runxin Xu, Qihao Zhu, Shirong Ma, Peiyi Wang, Xiao Bi, Xiaokang Zhang, Xingkai Yu, Yu~Wu, Z.~F. Wu, Zhibin Gou, Zhihong Shao, Zhuoshu Li, Ziyi Gao, Aixin Liu, and 180 others. 2025.
\newblock \href {https://arxiv.org/abs/2501.12948} {Deepseek-r1: Incentivizing reasoning capability in {LLM}s via reinforcement learning}.
\newblock \emph{Preprint}, arXiv:2501.12948.

\bibitem[{Huang et~al.(2025)Huang, Lai, Zhao, Xia, Bai, Sun, Liu, Liu, Pan, Tian et~al.}]{huang2025large}
Jiajie Huang, Honghao Lai, Weilong Zhao, Danni Xia, Chunyang Bai, Mingyao Sun, Jianing Liu, Jiayi Liu, Bei Pan, Jinhui Tian, and 1 others. 2025.
\newblock Large language model--assisted risk-of-bias assessment in randomized controlled trials using the revised risk-of-bias tool: Usability study.
\newblock \emph{Journal of Medical Internet Research}, 27:e70450.

\bibitem[{{Hugging Face}(2025)}]{openr1}
{Hugging Face}. 2025.
\newblock \href {https://github.com/huggingface/open-r1} {Open {R}1: A fully open reproduction of deepseek-r1}.

\bibitem[{Ji et~al.(2025)Ji, Zhao, Wang, Wang, Zhang, Cheng, Feng, and Zhang}]{ji2025robguard}
Changkai Ji, Bowen Zhao, Zhuoyao Wang, Yingwen Wang, Yuejie Zhang, Ying Cheng, Rui Feng, and Xiaobo Zhang. 2025.
\newblock Robguard: Enhancing llms to assess risk of bias in clinical trial documents.
\newblock In \emph{Proceedings of the 31st International Conference on Computational Linguistics}, pages 1258--1277.

\bibitem[{Kwon et~al.(2023)Kwon, Li, Zhuang, Sheng, Zheng, Yu, Gonzalez, Zhang, and Stoica}]{vllm}
Woosuk Kwon, Zhuohan Li, Siyuan Zhuang, Ying Sheng, Lianmin Zheng, Cody~Hao Yu, Joseph Gonzalez, Hao Zhang, and Ion Stoica. 2023.
\newblock Efficient memory management for large language model serving with pagedattention.
\newblock In \emph{Proceedings of the 29th symposium on operating systems principles}, pages 611--626.

\bibitem[{Lai et~al.(2025)Lai, Liu, Bai, Liu, Pan, Luo, Hou, Zhao, Xia, Tian et~al.}]{lai2025language}
Honghao Lai, Jiayi Liu, Chunyang Bai, Hui Liu, Bei Pan, Xufei Luo, Liangying Hou, Weilong Zhao, Danni Xia, Jinhui Tian, and 1 others. 2025.
\newblock Language models for data extraction and risk of bias assessment in complementary medicine.
\newblock \emph{npj Digital Medicine}, 8(1):74.

\bibitem[{Lambert et~al.(2025)Lambert, Morrison, Pyatkin, Huang, Ivison, Brahman, Miranda, Liu, Dziri, Lyu, Gu, Malik, Graf, Hwang, Yang, Bras, Tafjord, Wilhelm, Soldaini, Smith, Wang, Dasigi, and Hajishirzi}]{lambert2025tulu}
Nathan Lambert, Jacob Morrison, Valentina Pyatkin, Shengyi Huang, Hamish Ivison, Faeze Brahman, Lester James~Validad Miranda, Alisa Liu, Nouha Dziri, Xinxi Lyu, Yuling Gu, Saumya Malik, Victoria Graf, Jena~D. Hwang, Jiangjiang Yang, Ronan~Le Bras, Oyvind Tafjord, Christopher Wilhelm, Luca Soldaini, and 4 others. 2025.
\newblock \href {https://openreview.net/forum?id=i1uGbfHHpH} {Tulu 3: Pushing frontiers in open language model post-training}.
\newblock In \emph{Second Conference on Language Modeling}.

\bibitem[{Lightman et~al.(2024)Lightman, Kosaraju, Burda, Edwards, Baker, Lee, Leike, Schulman, Sutskever, and Cobbe}]{lightman2024lets}
Hunter Lightman, Vineet Kosaraju, Yuri Burda, Harrison Edwards, Bowen Baker, Teddy Lee, Jan Leike, John Schulman, Ilya Sutskever, and Karl Cobbe. 2024.
\newblock \href {https://openreview.net/forum?id=v8L0pN6EOi} {Let's verify step by step}.
\newblock In \emph{The Twelfth International Conference on Learning Representations}.

\bibitem[{Loshchilov and Hutter(2017)}]{adamw}
Ilya Loshchilov and Frank Hutter. 2017.
\newblock \href {https://api.semanticscholar.org/CorpusID:53592270} {Decoupled weight decay regularization}.
\newblock In \emph{International Conference on Learning Representations}.

\bibitem[{{OpenAI}(2024)}]{openai_o1_2024}
{OpenAI}. 2024.
\newblock Introducing openai o1.
\newblock \url{https://openai.com/it-IT/o1/}.

\bibitem[{{OpenAI}(2025)}]{openai_gpt5_2025}
{OpenAI}. 2025.
\newblock Introducing gpt-5.
\newblock \url{https://openai.com/it-IT/gpt-5/}.

\bibitem[{OpenAI et~al.(2025)OpenAI, Agarwal, Ahmad, Ai, Altman, Applebaum, Arbus, Arora, Bai, Baker, Bao, Barak, Bennett, Bertao, Brett, Brevdo, Brockman, Bubeck, Chang, Chen, Chen, Cheung, Clark, Cook, Dukhan, Dvorak, Fives, Fomenko, Garipov, Georgiev, Glaese, Gogineni, Goucher, Gross, Guzman, Hallman, Hehir, Heidecke, Helyar, Hu, Huet, Huh, Jain, Johnson, Koch, Kofman, Kundel, Kwon, Kyrylov, Le, Leclerc, Lennon, Lessans, Lezcano-Casado, Li, Li, Lin, Liss, Lily, Liu, Liu, Lu, Lu, Martinovic, McCallum, McGrath, McKinney, McLaughlin, Mei, Mostovoy, Mu, Myles, Neitz, Nichol, Pachocki, Paino, Palmie, Pantuliano, Parascandolo, Park, Pathak, Paz, Peran, Pimenov, Pokrass, Proehl, Qiu, Raila, Raso, Ren, Richardson, Robinson, Rotsted, Salman, Sanjeev, Schwarzer, Sculley, Sikchi, Simon, Singhal, Song, Stuckey, Sun, Tillet, Toizer, Tsimpourlas, Vyas, Wallace, Wang, Wang, Watkins, Weil, Wendling, Whinnery, Whitney, Wong, Yang, Yang, Yasunaga, Ying, Zaremba, Zhan, Zhang, Zhang, Zhang, and
  Zhao}]{openai2025gptoss120bgptoss20bmodel}
OpenAI, Sandhini Agarwal, Lama Ahmad, Jason Ai, Sam Altman, Andy Applebaum, Edwin Arbus, Rahul~K. Arora, Yu~Bai, Bowen Baker, Haiming Bao, Boaz Barak, Ally Bennett, Tyler Bertao, Nivedita Brett, Eugene Brevdo, Greg Brockman, Sebastien Bubeck, Che Chang, and 107 others. 2025.
\newblock \href {https://arxiv.org/abs/2508.10925} {gpt-oss-120b \& gpt-oss-20b model card}.
\newblock \emph{Preprint}, arXiv:2508.10925.

\bibitem[{Pronesti et~al.(2025{\natexlab{a}})Pronesti, Bettencourt-Silva, Flanagan, Pascale, Redmond, Belz, and Hou}]{pronesti-etal-2025-query}
Massimiliano Pronesti, Joao~H Bettencourt-Silva, Paul Flanagan, Alessandra Pascale, Ois{\'i}n Redmond, Anya Belz, and Yufang Hou. 2025{\natexlab{a}}.
\newblock \href {https://doi.org/10.18653/v1/2025.acl-long.1359} {Query-driven document-level scientific evidence extraction from biomedical studies}.
\newblock In \emph{Proceedings of the 63rd Annual Meeting of the Association for Computational Linguistics (Volume 1: Long Papers)}, pages 28034--28051, Vienna, Austria. Association for Computational Linguistics.

\bibitem[{Pronesti et~al.(2025{\natexlab{b}})Pronesti, Lorandi, Flanagan, Redmond, Belz, and Hou}]{pronesti-etal-2025-enhancing}
Massimiliano Pronesti, Michela Lorandi, Paul Flanagan, Ois{\'i}n Redmond, Anya Belz, and Yufang Hou. 2025{\natexlab{b}}.
\newblock \href {https://doi.org/10.18653/v1/2025.emnlp-main.1544} {Enhancing study-level inference from clinical trial papers via reinforcement learning-based numeric reasoning}.
\newblock In \emph{Proceedings of the 2025 Conference on Empirical Methods in Natural Language Processing}, pages 30345--30361, Suzhou, China. Association for Computational Linguistics.

\bibitem[{Shao et~al.(2024)Shao, Wang, Zhu, Xu, Song, Bi, Zhang, Zhang, Li, Wu, and Guo}]{deepseek-math}
Zhihong Shao, Peiyi Wang, Qihao Zhu, Runxin Xu, Junxiao Song, Xiao Bi, Haowei Zhang, Mingchuan Zhang, Y.~K. Li, Y.~Wu, and Daya Guo. 2024.
\newblock \href {https://arxiv.org/abs/2402.03300} {{D}eep{S}eek{M}ath: Pushing the limits of mathematical reasoning in open language models}.
\newblock \emph{Preprint}, arXiv:2402.03300.

\bibitem[{Skalse et~al.(2022)Skalse, Howe, Krasheninnikov, and Krueger}]{skalse2022defining}
Joar Skalse, Nikolaus Howe, Dmitrii Krasheninnikov, and David Krueger. 2022.
\newblock Defining and characterizing reward gaming.
\newblock \emph{Advances in Neural Information Processing Systems}, 35:9460--9471.

\bibitem[{Song et~al.(2025)Song, Su, Qu, Zhou, and Cheng}]{song-etal-2025-prmbench}
Mingyang Song, Zhaochen Su, Xiaoye Qu, Jiawei Zhou, and Yu~Cheng. 2025.
\newblock \href {https://doi.org/10.18653/v1/2025.acl-long.1230} {{PRMB}ench: A fine-grained and challenging benchmark for process-level reward models}.
\newblock In \emph{Proceedings of the 63rd Annual Meeting of the Association for Computational Linguistics (Volume 1: Long Papers)}, pages 25299--25346, Vienna, Austria. Association for Computational Linguistics.

\bibitem[{Sterne et~al.(2019)Sterne, Savović, Page, Elbers, Blencowe, Boutron, Cates, Cheng, Corbett, Eldridge, Hern{\'a}n, Hopewell, Hróbjartsson, Junqueira, Jüni, Kirkham, Lasserson, Li, McAleenan, Reeves, Shepperd, Shrier, Stewart, Tilling, White, Whiting, and Higgins}]{sterne2019rob2}
Jonathan~AC Sterne, Jelena Savović, Matthew~J. Page, Roy~G. Elbers, Natalie~S. Blencowe, Isabelle Boutron, Christopher~J. Cates, He~Cheng, Mark~S. Corbett, Sandra~M. Eldridge, Miguel~A. Hern{\'a}n, Sally Hopewell, Asbjørn Hróbjartsson, Diana~R. Junqueira, Peter Jüni, Jamie~J. Kirkham, Toby Lasserson, Tianjing Li, Ann McAleenan, and 8 others. 2019.
\newblock Ro{B} 2: a revised tool for assessing risk of bias in randomised trials.
\newblock \emph{BMJ}, 366:l4898.

\bibitem[{{\v{S}}uster et~al.(2024){\v{S}}uster, Baldwin, and Verspoor}]{vsuster2024zero}
Simon {\v{S}}uster, Timothy Baldwin, and Karin Verspoor. 2024.
\newblock Zero-and few-shot prompting of generative large language models provides weak assessment of risk of bias in clinical trials.
\newblock \emph{Research Synthesis Methods}, 15(6):988--1000.

\bibitem[{Wang et~al.(2025{\natexlab{a}})Wang, Cao, Bao, Zheng, Pasternak, Wang, Wang, Paturi, and Bergen}]{wang-etal-2025-measuring}
Jianyou Wang, Weili Cao, Longtian Bao, Youze Zheng, Gil Pasternak, Kaicheng Wang, Xiaoyue Wang, Ramamohan Paturi, and Leon Bergen. 2025{\natexlab{a}}.
\newblock \href {https://doi.org/10.18653/v1/2025.emnlp-main.160} {Measuring risk of bias in biomedical reports: The {R}o{BBR} benchmark}.
\newblock In \emph{Proceedings of the 2025 Conference on Empirical Methods in Natural Language Processing}, pages 3220--3248, Suzhou, China. Association for Computational Linguistics.

\bibitem[{Wang et~al.(2025{\natexlab{b}})Wang, Yang, Zeng, Ren, Liu, Peng, Cheng, He, Wang, Gao, Chen, Wang, Du, and yelong shen}]{wang2025reinforcement}
Yiping Wang, Qing Yang, Zhiyuan Zeng, Liliang Ren, Liyuan Liu, Baolin Peng, Hao Cheng, Xuehai He, Kuan Wang, Jianfeng Gao, Weizhu Chen, Shuohang Wang, Simon~Shaolei Du, and yelong shen. 2025{\natexlab{b}}.
\newblock \href {https://openreview.net/forum?id=IBrRNLr6JA} {Reinforcement learning for reasoning in large language models with one training example}.
\newblock In \emph{The Thirty-ninth Annual Conference on Neural Information Processing Systems}.

\bibitem[{Wang et~al.(2025{\natexlab{c}})Wang, Gao, and Ke}]{wang2025masked}
Zhen Wang, Zhifeng Gao, and Guolin Ke. 2025{\natexlab{c}}.
\newblock \href {https://arxiv.org/abs/2511.17473} {Masked-and-reordered self-supervision for reinforcement learning from verifiable rewards}.
\newblock \emph{arXiv preprint arXiv:2511.17473}.

\bibitem[{Wen et~al.(2025)Wen, Liu, Zheng, Xu, Ye, Wu et~al.}]{rlvr-incentivizes}
Xumeng Wen, Zihan Liu, Shun Zheng, Zhijian Xu, Shengyu Ye, Zhirong Wu, and 1 others. 2025.
\newblock Reinforcement learning with verifiable rewards implicitly incentivizes correct reasoning in base llms.
\newblock \emph{arXiv preprint arXiv:2506.14245}.

\bibitem[{Yang et~al.(2025{\natexlab{a}})Yang, Li, Yang, Zhang, Hui, Zheng, Yu, Gao, Huang, Lv, Zheng, Liu, Zhou, Huang, Hu, Ge, Wei, Lin, Tang, Yang, Tu, Zhang, Yang, Yang, Zhou, Zhou, Lin, Dang, Bao, Yang, Yu, Deng, Li, Xue, Li, Zhang, Wang, Zhu, Men, Gao, Liu, Luo, Li, Tang, Yin, Ren, Wang, Zhang, Ren, Fan, Su, Zhang, Zhang, Wan, Liu, Wang, Cui, Zhang, Zhou, and Qiu}]{yang2025qwen3technicalreport}
An~Yang, Anfeng Li, Baosong Yang, Beichen Zhang, Binyuan Hui, Bo~Zheng, Bowen Yu, Chang Gao, Chengen Huang, Chenxu Lv, Chujie Zheng, Dayiheng Liu, Fan Zhou, Fei Huang, Feng Hu, Hao Ge, Haoran Wei, Huan Lin, Jialong Tang, and 41 others. 2025{\natexlab{a}}.
\newblock \href {https://arxiv.org/abs/2505.09388} {Qwen3 technical report}.
\newblock \emph{Preprint}, arXiv:2505.09388.

\bibitem[{Yang et~al.(2025{\natexlab{b}})Yang, Yang, Zhang, Hui, Zheng, Yu, Li, Liu, Huang, Wei, Lin, Yang, Tu, Zhang, Yang, Yang, Zhou, Lin, Dang, Lu, Bao, Yang, Yu, Li, Xue, Zhang, Zhu, Men, Lin, Li, Tang, Xia, Ren, Ren, Fan, Su, Zhang, Wan, Liu, Cui, Zhang, and Qiu}]{qwen2.5}
An~Yang, Baosong Yang, Beichen Zhang, Binyuan Hui, Bo~Zheng, Bowen Yu, Chengyuan Li, Dayiheng Liu, Fei Huang, Haoran Wei, Huan Lin, Jian Yang, Jianhong Tu, Jianwei Zhang, Jianxin Yang, Jiaxi Yang, Jingren Zhou, Junyang Lin, Kai Dang, and 23 others. 2025{\natexlab{b}}.
\newblock \href {https://arxiv.org/abs/2412.15115} {Qwen2.5 technical report}.
\newblock \emph{Preprint}, arXiv:2412.15115.

\bibitem[{Yu et~al.(2025)Yu, Zhang, Zhu, Yuan, Zuo, YuYue, Dai, Fan, Liu, Liu, Liu, Liu, Lin, Lin, Ma, Sheng, Tong, Zhang, Zhang, Zhang, Zhang, Zhu, Zhu, Chen, Chen, Wang, Yu, Song, Wei, Zhou, Liu, Ma, Zhang, Yan, Wu, and Wang}]{yu2025dapo}
Qiying Yu, Zheng Zhang, Ruofei Zhu, Yufeng Yuan, Xiaochen Zuo, YuYue, Weinan Dai, Tiantian Fan, Gaohong Liu, Juncai Liu, LingJun Liu, Xin Liu, Haibin Lin, Zhiqi Lin, Bole Ma, Guangming Sheng, Yuxuan Tong, Chi Zhang, Mofan Zhang, and 17 others. 2025.
\newblock \href {https://openreview.net/forum?id=2a36EMSSTp} {{DAPO}: An open-source {LLM} reinforcement learning system at scale}.
\newblock In \emph{The Thirty-ninth Annual Conference on Neural Information Processing Systems}.

\bibitem[{Yun et~al.(2025)Yun, Sohn, Park, Kim, Tang, Shao, Koo, Minhyeok, Chen, Gerstein, Moor, and Kang}]{medprm}
Jaehoon Yun, Jiwoong Sohn, Jungwoo Park, Hyunjae Kim, Xiangru Tang, Daniel Shao, Yong~Hoe Koo, Ko~Minhyeok, Qingyu Chen, Mark Gerstein, Michael Moor, and Jaewoo Kang. 2025.
\newblock \href {https://doi.org/10.18653/v1/2025.emnlp-main.837} {{M}ed-{PRM}: Medical reasoning models with stepwise, guideline-verified process rewards}.
\newblock In \emph{Proceedings of the 2025 Conference on Empirical Methods in Natural Language Processing}, pages 16565--16582, Suzhou, China. Association for Computational Linguistics.

\bibitem[{Zelikman et~al.(2022)Zelikman, Wu, Mu, and Goodman}]{zelikman2022star}
Eric Zelikman, Yuhuai Wu, Jesse Mu, and Noah Goodman. 2022.
\newblock \href {https://openreview.net/forum?id=_3ELRdg2sgI} {{ST}ar: Bootstrapping reasoning with reasoning}.
\newblock In \emph{Advances in Neural Information Processing Systems}.

\bibitem[{Zeng et~al.(2025)Zeng, Huang, Liu, Liu, He, MA, and He}]{zeng2025simplerlzoo}
Weihao Zeng, Yuzhen Huang, Qian Liu, Wei Liu, Keqing He, Zejun MA, and Junxian He. 2025.
\newblock \href {https://openreview.net/forum?id=vSMCBUgrQj} {Simple{RL}-zoo: Investigating and taming zero reinforcement learning for open base models in the wild}.
\newblock In \emph{Second Conference on Language Modeling}.

\bibitem[{Zhang and Zuo(2025)}]{zhang-zuo-2025-grpo}
Jixiao Zhang and Chunsheng Zuo. 2025.
\newblock \href {https://doi.org/10.18653/v1/2025.emnlp-main.287} {{GRPO}-{LEAD}: A difficulty-aware reinforcement learning approach for concise mathematical reasoning in language models}.
\newblock In \emph{Proceedings of the 2025 Conference on Empirical Methods in Natural Language Processing}, pages 5642--5665, Suzhou, China. Association for Computational Linguistics.

\bibitem[{Zhang et~al.(2025)Zhang, Zheng, Wu, Zhang, Lin, Yu, Liu, Zhou, and Lin}]{zhang-etal-2025-lessons}
Zhenru Zhang, Chujie Zheng, Yangzhen Wu, Beichen Zhang, Runji Lin, Bowen Yu, Dayiheng Liu, Jingren Zhou, and Junyang Lin. 2025.
\newblock \href {https://doi.org/10.18653/v1/2025.findings-acl.547} {The lessons of developing process reward models in mathematical reasoning}.
\newblock In \emph{Findings of the Association for Computational Linguistics: ACL 2025}, pages 10495--10516, Vienna, Austria. Association for Computational Linguistics.

\bibitem[{Zou et~al.(2025)Zou, Yang, Gu, Qiu, Shen, He, and Wang}]{zou2025reasonfluxprm}
Jiaru Zou, Ling Yang, Jingwen Gu, Jiahao Qiu, Ke~Shen, Jingrui He, and Mengdi Wang. 2025.
\newblock \href {https://openreview.net/forum?id=f3sZjkQbv2} {Reasonflux-{PRM}: Trajectory-aware {PRM}s for long chain-of-thought reasoning in {LLM}s}.
\newblock In \emph{The Thirty-ninth Annual Conference on Neural Information Processing Systems}.

\end{thebibliography}

\appendix

\section{Theoretical Analysis}
\label{app:theo}

This section provides the proof of Theorem~1, which is an extension of prior results on
reinforcement learning with verifiable rewards (RLVR) in base LLMs~\citep{rlvr-incentivizes}.

\subsection{Setup and assumptions}
Let $R(Y)$ denote the verifiable process reward assigned to a response $Y$, and let $G$ be the number of responses sampled for the input $x$.  Let $\mathcal C$ denote the correctness event for a trajectory $Y$. Define
\begin{align*}
	\mu_c := \mathbb E[R(Y)\mid\mathcal C],\qquad &
	\mu_i := \mathbb E[R(Y)\mid\mathcal C^c],\\
	p := \mathbb P(\mathcal C),
\end{align*}

and let $m := p\mu_c + (1-p)\mu_i$ be the unconditional expected reward.

\paragraph{Assumption A1 (Verifiability and finite variance).}
For fixed $(x,r)$, the verifiable reward $R(Y)$ is a real-valued random
variable with finite mean and nonnegative variance $\sigma_Y>0$.

\paragraph{Assumption A2 (Reward gap).}
Correct reasoning trajectories have higher probabilities to induce correct answers:
$\mu_c>\mu_i$.

\paragraph{Assumption A3 (Concentration of group statistics).}
Let $\overline R$ and $S$ denote the empirical mean and standard deviation of
rewards within a sampled group. As the group size $G\to\infty$,

\[
\overline R \xrightarrow{p} m,
\qquad
S \xrightarrow{p} \sigma := \sqrt{\mathrm{Var}[R(Y)]}.
\]

\subsection{Normalised advantages}

GRPO uses the trajectory-level normalised advantage
\[
\hat A(Y) = \frac{R(Y)-\overline R}{S}.
\]
DAPO constructs token-level advantages by scaling the same trajectory advantage:
\[
\hat A_{i,t} = c_{i,t}\,\hat A(Y_i),
\]
where $c_{i,t}\ge0$ and $\frac{1}{\sum_i|Y_i|}\sum_{i,t}c_{i,t}=1$. Following~\citealp{rlvr-incentivizes} and without loss of generality, we consider a policy gradient update
\begin{align*}
\nabla J(\theta) \approx \cfrac{1}{G} \sum^G_{i=1} \hat{A}(Y_i) \nabla_\theta  \ log \ \pi_\theta (Y_i \mid x) .
\end{align*}
\subsection{Proof of Theorem 1}

We prove the result for $\hat A(Y)$; the DAPO case follows immediately by the nonnegativity of the constants $c_{i,t}$.

By Assumption A3 and Slutsky’s theorem,
\[
\hat A(Y)
= \frac{R(Y)-\overline R}{S}
\xrightarrow{d}
\frac{R(Y)-m}{\sigma}.
\]
Taking conditional expectations yields
\begin{align*}
\mathbb E[\hat A(Y)\mid\mathcal C]
\xrightarrow{G\to\infty}
\frac{\mu_c - m}{\sigma},
\\
\mathbb E[\hat A(Y)\mid\mathcal C^c]
\xrightarrow{G\to\infty}
\frac{\mu_i - m}{\sigma}.
\end{align*}

Substituting $m=p\mu_c+(1-p)\mu_i$ gives
\begin{align*}
\mu_c - m = (1-p)(\mu_c-\mu_i) > 0,
\\
\mu_i - m = -p(\mu_c-\mu_i) < 0,
\end{align*}

establishing the sign separation:
\begin{align*}
\mathbb E[\hat A(Y)\mid\mathcal C] > 0,
\\
\mathbb E[\hat A(Y)\mid\mathcal C^c] < 0.
\end{align*}

For DAPO,
\begin{align*}
\mathbb E[\hat A_{i,t}\mid\mathcal C]
= c_{i,t}\,\mathbb E[\hat A(Y)\mid\mathcal C] > 0,
\\
\mathbb E[\hat A_{i,t}\mid\mathcal C^c]
= c_{i,t}\,\mathbb E[\hat A(Y)\mid\mathcal C^c] < 0.
\end{align*}

Thus, both GRPO and DAPO apply positive expected weight to correct traces and
negative weight to incorrect ones, proving Theorem~1.

\subsection{Verifiable Outcome Reward as a Special Case of VPRMs}	Define the degenerate label spaces $\mathcal L_t^{(r)}=\varnothing$ for all $t<T$. Then $r_t(\cdot)=0$ for $t<T$ and
	\[
	R(Y;x,r)=r_{\mathrm{label}}(Y;x,r).
	\]
	Hence a Verifiable Outcome Reward Model is exactly the case of a VPRM with no intermediate verifiable labels. All the results above apply: they reduce to the original GRPO/DAPO statements where the scalar reward depends only on final outcome statistics.

\section{Risk of Bias Assessment}\label{app:rob}
Risk-of-bias estimation evaluates the extent to which study findings may be systematically distorted. The process is organised into a set of domains that correspond to common sources of bias in randomized trials. For each domain, reviewers extract relevant information from the study report and translate it into qualitative judgments about the presence and potential impact of bias. 
Modern assessment tools, such as RoB 2.0~\cite{sterne2019rob2}, increasingly leverage automated decision rules to standardise these judgments. Algorithm~\ref{alg:macroA} provides an example of a macro for risk of type A using the labels defined in this paper, which illustrates how extracted steps are mapped to specific risk levels. Below we outline the main domains considered in our work, together with the typical reasoning steps involved.

\paragraph{A. Random sequence generation}
This domain assesses whether the method used to generate the allocation sequence was truly random. Reviewers first check whether the study reports how randomization was carried out. If so, they evaluate the nature of the method (e.g., computer-generated sequence versus quasi-random methods such as alternation) and judge whether the sequence could have been predicted. Clearly reported and genuinely random procedures indicate low risk; quasi-random or non-random procedures, or a lack of information, increase concern.

\paragraph{B. Allocation concealment}
Here the question is whether the assignment to treatment groups was shielded from those enrolling participants. Reviewers determine whether concealment was reported and whether the method (e.g., sealed opaque envelopes, central allocation) prevented foreknowledge of upcoming assignments. Adequate concealment protects against selection bias, whereas inadequate or unclear procedures raise concerns.

\paragraph{C. Blinding of participants and personnel}
This domain considers whether participants and those administering interventions were aware of group assignments. Reviewers establish whether blinding was reported, whether it involved participants, personnel, or both, and whether the blinding approach was likely to have been effective. Lack of blinding, or ineffective procedures, may influence participants' behaviour or care delivery and thus introduce performance bias.

\paragraph{D. Blinding of outcome assessment}
Assessors may also be influenced by knowledge of treatment allocation. Reviewers check whether outcome assessors were blinded and whether blinding was likely to minimise biased measurement. Absence of blinding or unclear reporting raises the possibility that assessments were influenced by expectations or prior beliefs.

\paragraph{E. Incomplete outcome data}
This domain evaluates the extent and handling of missing data. Reviewers consider how much data is missing, whether reasons for missingness are reported and plausible, and whether the analysis appropriately accounts for missing data. High or unexplained attrition, or inadequate handling strategies, can produce biased estimates of effect.

\paragraph{F. Selective reporting}
Selective reporting bias arises when outcomes are reported inconsistently with the study protocol or when unplanned outcomes are introduced. Reviewers check whether a protocol is available, compare planned and reported outcomes, and assess whether omissions or additions suggest selective emphasis. Clear correspondence indicates low risk; discrepancies raise concern.

\medskip
\noindent In addition to these core domains, we also consider supplementary aspects relevant to internal validity: similarity of baseline outcomes (G), similarity of baseline characteristics (H), and risk of contamination between study arms (I). These domains capture further sources of potential bias arising from imbalances at baseline or from unintended exposure to interventions across groups.

\begin{algorithm}
    \captionsetup{size=scriptsize} 
    \scriptsize
    \begin{algorithmic}[1]
        \Procedure{PredictLabel-A}{steps}
        \If{steps[\textsc{IdentifyRandomizationReport}] = \textsc{NotReported}}
        \State \Return \textsc{Moderate}
        \EndIf
        \If{steps[\textsc{ClassifyRandomizationMethod}] = \textsc{NonRandom}}
        \State \Return \textsc{High}
        \EndIf
        \If{steps[\textsc{AssessSequencePredictability}] = \textsc{Predictable}}
        \State \Return \textsc{Moderate}
        \EndIf
        \If{steps[\textsc{BaselineImbalance}] = \textsc{Likely}}
        \State \Return \textsc{High}
        \EndIf
        \State \Return \textsc{Low}
        \EndProcedure
    \end{algorithmic}
    \caption{RoB A Macro}\label{alg:macroA}
\end{algorithm}

\begin{table}[t]
\resizebox{\columnwidth}{!}{
\centering
\begin{tabular}{lccccccccc}
\toprule
\textbf{Dataset} & \textbf{A} & \textbf{B} & \textbf{C} & \textbf{D} & \textbf{E} & \textbf{F} & \textbf{G} & \textbf{H} & \textbf{I}\\
\midrule
\textsc{CochraneForestExt} & 498 & 498 & 498 & 498 & 498 & 273 &  61 & 61 & 61 \\
\textsc{CochraneForest} & 330 & 330 & 330 & 330 & 330 & 112 & 28  & 28 & 28 \\
RoBBR Cochrane & 125 & 125 & 198 & 133 & 206 & 119 & 0 & 0 & 0 \\
RoBBR Non-Cochrane & 412 & 474 & 467 & 472 & 478 & 186 & 0 & 0  & 0 \\
\bottomrule
\end{tabular}
}
\caption{Datasets statistics per risk type.}
\label{tab:dataset_stats_per_risk}
\end{table}

\section{Prompts}\label{app:prompts}
The prompts used for synthetic data annotation and
for training are shown in Figure~\ref{fig:synth_prompt} and~\ref{fig:train_inf_prompt}, respectively. For training, a temperature of 0.7 and 2,048
tokens as maximum output length are used.

\begin{figure}
	\scriptsize
	\begin{tcolorbox}[
		colback=gray!10, 
		colframe=black, 
		boxrule=0.5mm,
		title={Prompt for synthetic data annotation}, 
		fonttitle=\bfseries 
		]
		Articles: \{articles\}
		\\
		\\
		Your task is to produce a structured reasoning trace for the following risk of bias domain to justify the ground truth value. \\
		\\	
		Comparison: \{comparison\}\\
		Outcome: \{outcome\} \\
		Bias:  \{bias\_id\} -- \{bias\_definition\} \\
		Ground\_truth: \{bias\_value\} \\
		\\
		You must follow the structured reasoning procedure defined for each risk-of-bias domain (A–I).	For every domain, you must use the exact step names and allowable categorical labels listed below:
		\\\\
		\{steps\_and\_labels\}
		\\
		\\
		Follow this exact output structure:\\
		\{\{\\
		"step\_name": "step\_label", \\
		"step\_name\_rationale": "your detailed rationale", \\
		... \\ 
		(repeat for all steps required by the bias domain)\\
		\}\}
	\end{tcolorbox}
	\caption{Prompt for synthetic data annotation.}
	\label{fig:synth_prompt}
\end{figure}

\begin{figure}
	\scriptsize
	\begin{tcolorbox}[
		colback=gray!10, 
		colframe=black, 
		boxrule=0.5mm,
		width=\linewidth,
		title={Steps and labels}, 
		fonttitle=\bfseries 
		]
	    A — Random sequence generation\\  
		Identify\_randomization\_report → reported | not\_reported \\ 
		Classify\_randomization\_method → random | non\_random  \\ 
		Assess\_sequence\_predictability → unpredictable | predictable\\
		Baseline\_imbalance → likely | none \\
		\\ 
		B — Allocation concealment  \\
		Identify\_concealment\_report → reported | not\_reported \\ 
		Determine\_concealment\_method → adequate | inadequate   \\
		Assess\_possibility\_of\_foreknowledge → no | possible  \\
		\\ 
		C — Blinding of participants and personnel  \\
		Identify\_blinding\_report → reported | not\_reported  \\
		Assess\_blinding\_status → participants | personnel | both | none   \\
		Evaluate\_blinding\_effectiveness → effective | ineffective  \\
		\\
		D — Blinding of outcome assessment  \\
		Identify\_outcome\_blinding\_report → reported | not\_reported \\
		Assess\_assessor\_blinding → yes | no   \\
		Evaluate\_blinding\_effect\_on\_measurement → no | possible \\
		\\
		E — Incomplete outcome data  \\
		Quantify\_missing\_data → none | low | high   \\
		Identify\_missing\_data\_reason → adequate | inadequate | not\_reported  \\
		Assess\_handling\_of\_missing\_data → appropriate | inappropriate   \\
		Estimate\_bias\_due\_to\_missing\_data → unlikely | likely  \\
		\\
		F — Selective reporting  \\
		Identify\_protocol\_availability → available | not\_available  \\
		Compare\_outcomes\_reported → all | partial | none   \\
		Detect\_unexpected\_outcomes → none | added   \\
		Evaluate\_reporting\_selectivity → no | possible | yes  \\
		\\
		G — Baseline outcomes similar  \\
		Identify\_baseline\_outcomes\_report → reported | not\_reported  \\
		Compare\_baseline\_outcomes → similar | different   \\
		Evaluate\_impact\_of\_differences → likely\_impact | unlikely\_impact  \\
		\\
		H — Baseline characteristics similar  \\ 
		Identify\_baseline\_characteristics\_report → reported | not\_reported  \\
		Compare\_baseline\_characteristics → similar | different   \\
		Evaluate\_impact\_of\_differences → likely\_impact | unlikely\_impact  \\
		\\
		I — Contamination  \\
		Identify\_contamination\_risk\_report → reported | not\_reported   \\
		Assess\_contamination\_possibility → possible | unlikely   \\
		Assess\_contamination\_impact → likely\_impact | unlikely\_impact  \\
	\end{tcolorbox}
\caption{Steps and labels.}
\label{fig:steps_labels}
\end{figure}

\begin{figure}
    \scriptsize
	\begin{tcolorbox}[
		colback=gray!10, 
		colframe=black, 
		boxrule=0.5mm,
		width=\linewidth,
		title={Prompt for training and inference}, 
		fonttitle=\bfseries 
		]
		Articles: \{articles\}
		\\
		\\
		Question: Based on the given article, what is the risk of bias for the following Comparison and Outcome?
		\\
		Comparison: \{comparison\}\\
		Outcome: \{outcome\} \\
		The bias you have to assess is defined as follows: \{bias\}
		\\ 
		You must follow the structured reasoning procedure defined for each risk-of-bias domain (A–I).	For every domain, you must use the exact step names and allowable categorical labels listed below:
		\\\\
        \{steps\_and\_labels\}
        \\
		\\
		Follow this exact structure for your reasoning:
		\\
		<think>\\
		Step 1: step\_name \\
		...your thought process here...\\
		Answer: step\_label\\
		\\
		Step 2: step\_name\\
		...your thought process here...\\
		Answer: step\_label\\
		\\
		(repeat for all steps required by the bias domain)\\
		</think>\\
		<answer>\\
		risk: high | low | moderate\\
		</answer>
	\end{tcolorbox}
	\caption{Prompt for VPRM-training and -inference.}
	\label{fig:train_inf_prompt}
\end{figure}

\section{Silver Steps and Labels Manual Verification}\label{app:silver_labels} 
To assess the quality of the automatically generated reasoning steps and silver labels used for VPRM training, we selected a random sample of 20 instances from the full dataset and manually evaluated their correctness. The evaluation was conducted by two master’s students in NLP familiar with the task. For each instance, annotators inspected the complete step-level reasoning trace and verified two properties: (i) whether the sequence of steps constituted a valid decision path for the target risk-of-bias domain, according to Cochrane’s domain-specific guidelines; and (ii) whether each step and label was valid given the underlying paper. For each item, we recorded whether the overall reasoning trace was coherent, and for each individual step we recorded whether the step and its label were valid.

\begin{table}[h]
\small
\centering
\begin{tabular}{l c}
\toprule
\textbf{Metric} & \textbf{Fraction} \\
\midrule
Coherent instances & 100.0 \% \\
Correct steps & 100.0 \% \\
Correct labels & 96.7\% \\
\bottomrule
\end{tabular}
\caption{Manual verification of 20 randomly sampled silver-labelled reasoning traces.}
\label{tab:silver_verification}
\end{table}

Table~\ref{tab:silver_verification} summarises the proportion of coherent instances, correct steps, and correct labels observed in this manual evaluation. Results of the manual verification show that the automatically generated steps and labels used for VPRM training are of consistently high quality. All inspected traces follow the correct decision structure, and step-level labels are almost always accurate, providing a solid ground for model training.

\section{Related Work}\label{app:related_work}
\paragraph{Reinforcement Learning and Verifiable Rewards} Several extensions of reinforcement learning with verifiable rewards (RLVR) go beyond outcome-only supervision by enriching the reward signal with structural information. Masked-and-reordered self-supervision provides auxiliary signals encouraging coherent intermediate reasoning~\cite{wang2025masked}, while self-verification methods add progress-estimation or critique modules that guide models toward more reliable reasoning trajectories~\cite{zeng2025simplerlzoo}. Theoretical analyses further show that verifiable rewards influence trajectory selection in predictable ways, steering models toward high-success modes under verifiable criteria~\cite{rlvr-incentivizes}. These approaches strengthen RLVR but remain fundamentally centered on terminal-outcome verification.

Complementary lines of research pursue process supervision, scoring CoT steps using neural judges~\cite{lightman2024lets, zelikman2022star, zhang-etal-2025-lessons, zou2025reasonfluxprm}. While such methods provide dense feedback unavailable to outcome-only RL, they depend on model-generated evaluations and therefore inherit issues of opacity, bias, and reward hacking~\cite{amodei2016concrete, skalse2022defining}. Crucially, their intermediate rewards are not verifiable.

Taken together, these works highlight two remaining gaps: existing approaches lack (i) \emph{verifiability} of intermediate rewards and (ii) \emph{fine-grained} step-level supervision grounded in deterministic rules. To date, no method provides reinforcement learning over reasoning trajectories where every step is evaluated by an externally checkable verifier. Verifiable Process Reward Models (VPRMs) address this gap by combining the robustness of RLVR with step-wise, rule-based verification, which not only enables transparent, structurally aligned reasoning but also removes the opportunities for reward hacking inherent in neural process rewards.

\paragraph{RoB Assessment and Automated Evidence Evaluation.} 
Prior work on automated RoB assessment has largely relied on supervised modelling or prompted LLMs. 
Transformer-based systems such as RoBIn~\cite{dias2025robin} frame RoB inference as a machine reading comprehension task and train classifiers directly on annotated evidence. Other approaches enhance pretrained LLMs with retrieval or auxiliary decision heads, as in RoBGuard~\cite{ji2025robguard}.
Several studies investigate LLM prompting for RoB assessment, reporting limited reliability when models operate without explicit procedural constraints~\cite{huang2025large,vsuster2024zero}. Likewise, analyses of LLM-based critical appraisal highlight dependence on model pretraining and prompt sensitivity rather than verifiable optimisation~\cite{wang-etal-2025-measuring,lai2025language}. 
Across these methodologies, existing systems employ prompting or supervised fine-tuning, but none leverage reinforcement learning for RoB assessment. 
Our work is, to our knowledge, the first to introduce RL-based training in this domain.

\section{Hyperparameters and APIs}\label{app:hyp_and_apis} We executed all the experiments either via API or on our own cluster. We used the paid-for OpenAI API to access GPT-4. On the other hand, we hosted and trained the open-source models used in this paper on a distributed cluster. 

SFT is performed for 5 epochs with a batch size of 1 (due to the large size of the input data) using
a learning rate of $5 \times 10^{-5}$ and the AdamW optimiser~\cite{adamw}. 

For the RL setups, we adopt the GRPO~\cite{deepseek-math} and DAPO~\cite{yu2025dapo} algorithms, training for 3 epochs with a learning rate of $1 \times 10^{-6}$, per-device batch size 1, and 16 sampled generations per batch. Both training protocols leverage gradient accumulation with 8 accumulation steps. All experiments are conducted using the Open-R1
framework~\cite{openr1} on 8 NVIDIA A100 GPUs, each equipped with 80GB of memory. Models have been served for inference with the vLLM framework~\cite{vllm}.

\section{Scientific Artefacts and Licensing} In this work, we used the following scientific artefacts. LLaMa 3.1 is licensed under a commercial license\footnote{\url{https://llama.meta.com/doc/overview}}. GPT-4 is licensed under a commercial license\footnote{\url{https://openai.com/policies/terms-of-use}}. Qwen2.5 is licensed under the Apache 2.0 license\footnote{\url{https://qwenlm.github.io/blog/qwen3}}. Granite 3.1 is licensed under the Apache 2.0 license\footnote{\url{https://www.ibm.com/architectures/product-guides/granite-31}}.  DeepSeek models are licensed under the MIT license\footnote{\url{https://api-docs.deepseek.com/news/news250120}}. Mining text and data from the Cochrane library is permitted for non-commercial research through the Wiley API.\footnote{\url{https://www.cochranelibrary.com/help/access}}. The usage of the listed artefacts is consistent with their licenses.

\begin{figure*}
    \centering
    \includegraphics[width=0.9\linewidth]{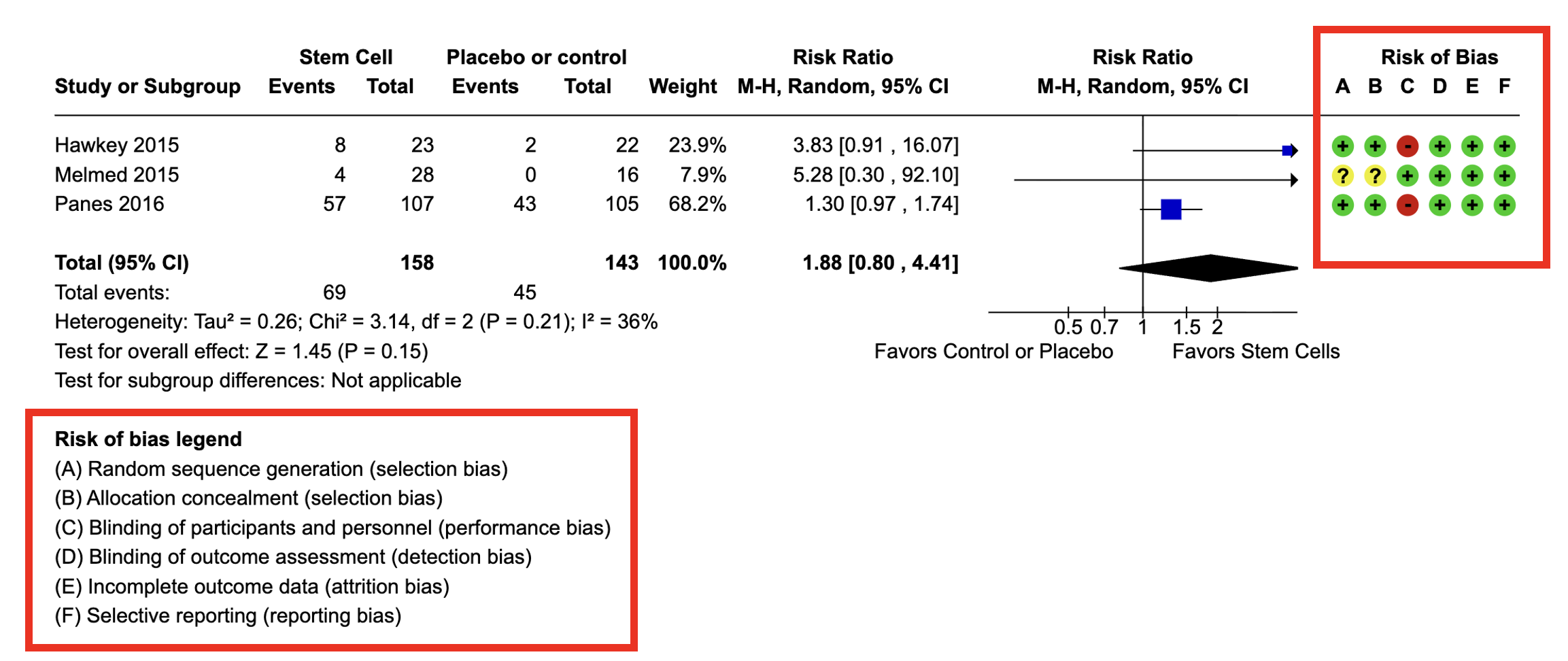}
    \caption{Example of forest plot containing three studies and their corresponding risk of bias map.}
    \label{fig:example_fp}
\end{figure*}

 \begin{table*}[h!]
    \small
	     \centering
	     \begin{tabular}{l|>{\raggedright\arraybackslash}p{0.7\textwidth}}
		     \toprule
		     \rowcolor[gray]{0.9} \multicolumn{2}{c}{\textbf{Hawkey 2015}} \\
		     \midrule
		         \textbf{Text} & Autologous Hematopoetic Stem Cell Transplantation for Refractory Crohn Disease: A Randomised Clinical Trial
                 \newline\newline
                 \textbf{Importance}  Case reports and series suggest hematopoietic stem cell transplantation (HSCT) may benefit some patients with Crohn disease. \newline ... \\ 
		         \midrule
		         \textbf{Outcome} &  Clinical Remission \\
              \textbf{Comparison} & Stem Cell vs Placebo or control \\
		       \textbf{Bias type}  & A \\
              \textbf{Bias name} & Random sequence generation (selection bias) \\
              \textbf{Bias definition} & Bias arising from the randomization process\\
              \textbf{Bias value} & \textbf{+} \\ 
                 \midrule
                \textbf{Annotation} & 
                \begin{minipage}[t]
                {0.7\textwidth}
                \footnotesize
                \raggedright  
                "Identify\_randomization\_report": "reported",\\
                "Identify\_randomization\_report\_rationale": "The methods section of the trial explicitly states that participants were allocated to groups using a computer-generated random sequence. The authors describe the randomisation process in sufficient detail, indicating that a statistician who was not involved in participant recruitment generated the sequence.",\\[0.5em]
                "Classify\_randomization\_method": "random",\\
                "Classify\_randomization\_method\_rationale": "A computer-generated sequence (e.g., using a random number generator or statistical software) is a truly random method. It does not rely on any deterministic or quasi-random technique such as alternation, date of birth, or admission order, therefore it is classified as 'random'.",\\[0.5em]
                "Assess\_sequence\_predictability": "unpredictable",\\
                "Assess\_sequence\_predictability\_rationale": "Because the sequence was generated by a computer algorithm and concealed from the investigators enrolling participants (allocation was performed via sealed, opaque envelopes prepared after the sequence was generated), neither the investigators nor the participants could foresee the upcoming allocation. This makes the sequence unpredictable.",\\[0.5em]
                "Baseline\_Imbalance": "none",\\
                "Baseline\_Imbalance\_rationale": "Baseline tables show that demographic and clinical characteristics (age, sex, disease severity, etc.) are comparable across groups with no statistically significant differences, indicating that the randomisation produced balanced groups and is unlikely to have introduced bias."
          \end{minipage} \\
		     \bottomrule
		     \end{tabular}
	     \caption{The corresponding entry in \textsc{CochraneForestExt} for Hawkey 2015 in Figure~\ref{fig:example_fp}. Rationales are not used for RL training.}
	     \label{tab:data_example}
	 \end{table*}
     
\end{document}